\documentclass[conference]{IEEEtran}
\usepackage{times}

\usepackage[numbers]{natbib}
\usepackage{multicol}

\usepackage{array}

\usepackage{epsfig}
\usepackage{graphicx}
\usepackage{booktabs}
\usepackage{multirow}
\usepackage{bigstrut}
\usepackage{mathptmx} 
\usepackage{times} 
\usepackage{amsmath} 
\usepackage{amssymb}  
\usepackage{amsmath,amssymb,amsfonts}
\usepackage{textcomp}
\usepackage{algorithm}
\usepackage{algpseudocode}
\usepackage{cite}
\usepackage{multirow}

\usepackage[bookmarks=true]{hyperref}

\begin{document}

\title{Sequential Place Learning: Heuristic-Free High-Performance Long-Term Place Recognition}



\author{Marvin Chanc\'an, Michael Milford}



%

\maketitle
\begin{abstract}
Sequential matching using hand-crafted heuristics has been standard practice in route-based place recognition for enhancing pairwise similarity results for nearly a decade. However, precision-recall performance of these algorithms dramatically degrades when searching on short \textit{temporal window} (TW) lengths, while demanding high compute and storage costs on large robotic datasets for autonomous navigation research. Here, influenced by biological systems that robustly navigate spacetime scales even without vision, we develop a \textit{joint visual and positional representation learning} technique, via a \textit{sequential process}, and design a learning-based CNN+LSTM architecture, trainable via backpropagation through time, for viewpoint- and appearance-invariant place recognition. Our approach, \textit{Sequential Place Learning} (SPL), is based on a CNN function that visually encodes an environment from a single traversal, thus reducing storage capacity, while an LSTM temporally fuses each visual embedding with corresponding positional data---obtained from any source of motion estimation---for direct sequential inference. Contrary to classical two-stage pipelines, \textit{e.g.}, \textit{match-then-temporally-filter}, our network directly eliminates false-positive rates while jointly learning sequence matching from a single monocular image sequence, even using short TWs. Hence, we demonstrate that our model outperforms 15 classical methods while setting \textit{new state-of-the-art} performance standards on 4 challenging benchmark datasets, where one of them can be considered solved with recall rates of 100\% at 100\% precision, correctly matching \textit{all} places under extreme sunlight-darkness changes. In addition, we show that SPL can be up to 70$\times$ faster to deploy than classical methods on a 729 km route comprising 35,768 consecutive frames. Extensive experiments demonstrate the potential of this framework through quantitative and qualitative results. Baseline code available at \url{https://github.com/mchancan/deepseqslam}
\end{abstract}
\IEEEpeerreviewmaketitle

\section{Introduction}

The vast majority of sequential filtering algorithms in place recognition research for autonomous robot navigation are variants of carefully designed, complex hand-crafted heuristics. The use of these algorithms on top of pre-computed global image descriptors based on convolutional neural networks (CNN) \citep{deeplearning} has enabled researchers to improve significantly accuracies on challenging driving datasets under viewpoint and appearance changes, \textit{e.g.}, different weather, season or illumination \citep{8756053}. The success of these multi-frame filtering methods primarily rely on leveraging temporal information---explicitly found in image sequences recorded from large outdoor real environments---for reducing high false-positive rates typically found in sequence-based place recognition tasks due to perceptual aliasing and environmental cycles \citep{vprsurvey}.

\begin{figure}[!ht]
  \centering
  \includegraphics[width=\linewidth]{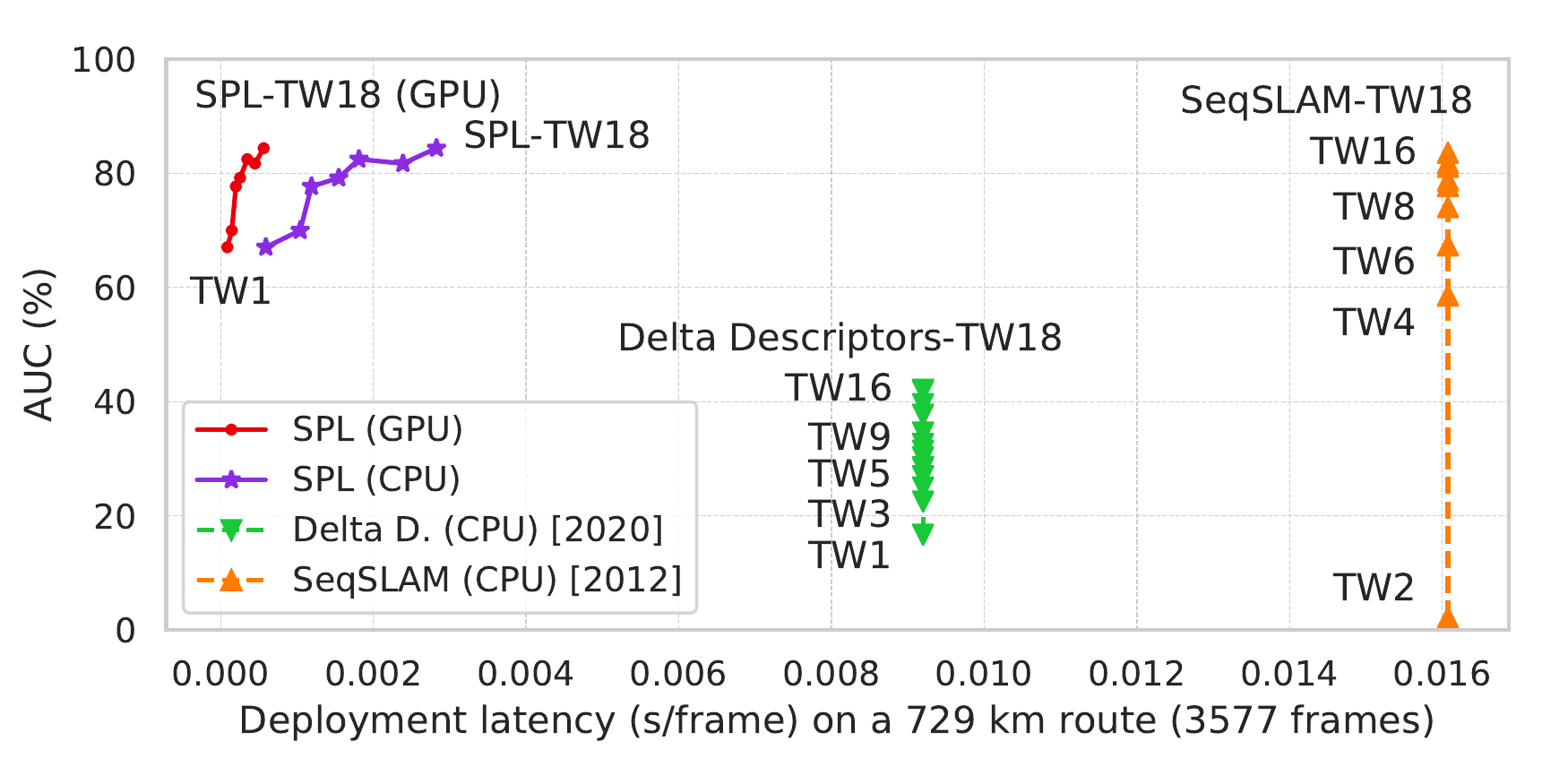}
  \vspace{-6mm}
  \caption{\textbf{Average Precision vs. Testing Latency}. For testing, we sampled 3K+ frames of the 729 km Nordland dataset for better visualization of deployment latencies. All our models achieve comparable results to the best SeqSLAM settings, even using short TW, while being $\sim10\times$ faster to deploy. Our approach can be $\sim60\times$ faster than classical methods on the full 35K+ frames. \textit{Note:} Speed-up values are based on average latencies of our model running on CPU, as classical methods can be deployed on CPU only.}
  \vspace{-6mm}
  \label{latency}
\end{figure}

However, heuristic-based sequential matching has three significant practical disadvantages. First, its computational cost scales with the dataset size, which incurs expensive memory overhead, and increases the time required to deploy on big datasets or autonomous driving applications. Second, its accuracy rapidly degrades when reducing key searching parameter values such as \textit{temporal window} (TW) lengths. Third, it typically requires both reference and query traversals during deployment, which increases storage cost with dataset size, to compute a pairwise similarity matrix that is then directly enhanced based on best-match searching heuristics or strong assumptions on velocity information for instance. Moreover, the development of entirely new multi-frame filtering methods, through the use of recurrent neural networks (RNNs) \citep{rnn}, suitable for learning structure in time, has been surprisingly limited; only recently the use of hand-crafted, non-trainable continuous attractor recurrent networks has been proposed \citep{chancan2020hybrid}.

In this paper, we argue that these limitations and negative consequences can be fundamentally addressed in a deep learning framework by exploiting the underinvestigated use of RNNs for sequence inference, rather than heuristic-based techniques. Hence, we build on using long short-term memory (LSTM) \citep{lstm} networks for learning sequence filtering tasks, addressing all the main limitations of classical methods. Our main contributions are summarized as follows:

\begin{itemize}
    \item We design a trainable CNN+LSTM architecture, which fuses \textit{visual and positional} data, recorded from a single traversal of an environment, via a \textit{sequential process}. Our approach, \textit{Sequential Place Learning} (SPL), learns multi-frame sequential matching tasks end-to-end via backpropagation through time (BPTT) \citep{WERBOS1988339}.
    \item We propose a sequence processing strategy that allows us to train our model to achieve substantially higher average-precision results using short TW lenghts, contrary to what was known to-date based on classical sequential filtering methods for place recognition in robotics research \citep{howlow}.
    \item We present the first demonstration of high-performance learning-based sequence filtering on a range of deployment latencies (see Fig. \ref{latency}), which allows us to set \textit{new state-of-the-art results} on \textit{4 benchmark robotic datasets}, while \textit{outperforming 15 classical methods}.
\end{itemize}



Sequential place learning (SPL) simultaneously train two key learning-based systems: (i) a convolutional neural network (CNN) for encoding raw RGB images, and (ii) a recurrent network (LSTM) for sequential learning and inference by temporally fusing visual encodings with positional data; obtained from any source of motion estimation including but not limited to wheel/visual/radar odometry, LiDAR, GPS, structure from motion (SfM) or even synthetic time-series data. 

We conduct extensive experiments to demonstrate the significance of SPL on four (4) datasets (Oxford RobotCar, Nordland Railway, St. Lucia, and Gardens Point) each recorded multiple times along 10-km, 729-km, 1-km and $\sim$10-km routes, respectively, with diverse conditions. On the Gardens Point dataset, in particular, we compare our model against ten (10) well-known classical methods under drastic day-night changes and show that our approach attains 100\% recall rates at 100\% precision for the first time.

The paper is structured as follows. We discuss the problem of multi-frame place recognition tasks while briefly surveying research papers on classical sequential filtering techniques in Section \ref{sec:vpr}. We introduce the concepts of \textit{simultaneous visual-and-positional learning } and \textit{multi-frame place learning} for route-based place recognition tasks, and describe how we developed our new state-of-the-art architectures in Section \ref{sec:plearning}. Finally, we present our experimental setup and results in Sections \ref{sec:setup} and \ref{sec:results}, respectively, before conclusion in Section \ref{sec:conclusion}; which is followed by the Appendix with additional results.

\section{Understanding Sequence Filtering}
\label{sec:vpr}

Here we first describe how pairwise (reference-query) image similarities work well for single-frame, image-retrieval-like tasks, but produce poor results when dealing with image sequences. Highlighting the need for versatile sequence filtering methods for route-based place recognition. We then briefly review prior work on sequence matching and describe the key components that might allow us to understand key strategies within classical methods in order to replicate these benefits when training neural networks for sequence modeling tasks.

\subsection{Image Similarity}



Visual place recognition research currently relies on image representations for global image description typically obtained from off-the-shelf CNN functions, rather than classical hand-crafted features such as SIFT \citep{lowe1999object} or SURF \citep{bay2006surf}, pre-trained for classification \citep{convnetlandmarks} or image retrieval tasks \citep{netvlad}.





\subsection{Sequence Matching}

The majority of multi-frame place recognition methods are then built based on pre-computed pairwise difference similarity matrices, between reference and query image representations, for iteratively applying temporal filtering/searching heuristics that seek to reduce false-positive rates such as ABLE \citep{arroyo2015towards}, ISM \citep{vysotska2015efficient}, OPR \citep{vysotska2015lazy}, VPR \citep{vysotska2017relocalization}, HMM \citep{hansen2014visual}, MCN \citep{8756053} and many others \citep{9109951,deltad,8756053, lowry2018lightweight,chen2014convolutional,avp}. The key ideas behind this, now standard, methodology date from key appearance-only topological place recognition systems such as FAB-MAP \citep{cummins2008fab, cummins2009highly} and SeqSLAM \citep{seqslam,howlow}.

Subsequent work that integrated odometry information was shown to improve place recognition performance such as CAT-SLAM \citep{maddern2012cat} or SMART \citep{smart} on a variety of benchmark datasets \citep{schubert2019towards}. The main benefits of the sequence filtering layers of these systems is that it allows robust generalization across challenging appearance and viewpoint variations including multiple seasons, weather, and lighting conditions \citep{chancan2020deepseqslam}. All these models, however, share a common limitation which is that they do not incorporate modern learning-based systems, \textit{e.g}, recurrent neural networks (RNN), for sequence inference.



\section{Towards Sequential Place Learning}
\label{sec:plearning}

Only recently, researchers have shown the use of dynamical attractors neural circuits \citep{miller2016dynamical} (a.k.a. continuous attractor neural networks (CANN)), a neuroscience-oriented type of RNN, to perform sequential place recognition without resorting to classical methods for sequence matching \citep{chancan2020hybrid}. This work demonstrate the potential of RNNs for temporal processing tasks in robotics research as in recent hybrid approaches \citep{zou2020hybrid,wang2021end}. Although the model proposed in \citep{chancan2020hybrid} achieves competitive results on challenging place recognition datasets, the CANN component does not incorporate learning capabilities, instead relying on pre-assigned unit interconnections and weights that need to be carefully fine-tuned to deploy on a particular dataset. Also, it does not incorporate self position-aware learning properties, rather it implements a direct mechanistic shift-and-copy action to simulate movement through the environment.

\subsection{Joint Visual-and-Positional Place Learning}


In contrast with classical two-stage multi-frame pipelines, \textit{e.g.}, \textit{match-then-temporally-filter}, here we propose to jointly learn visual-and-positional representations that can be simultaneously used for sequential inference for place recognition for the first time. We design and implement an entire neural network  which can be trained end-to-end via BPTT \citep{WERBOS1988339}. In the next section, we describe how the visual and positional processing components of our architecture are integrated.



\begin{figure}[!t]
  \centering
  \includegraphics[width=0.95\linewidth]{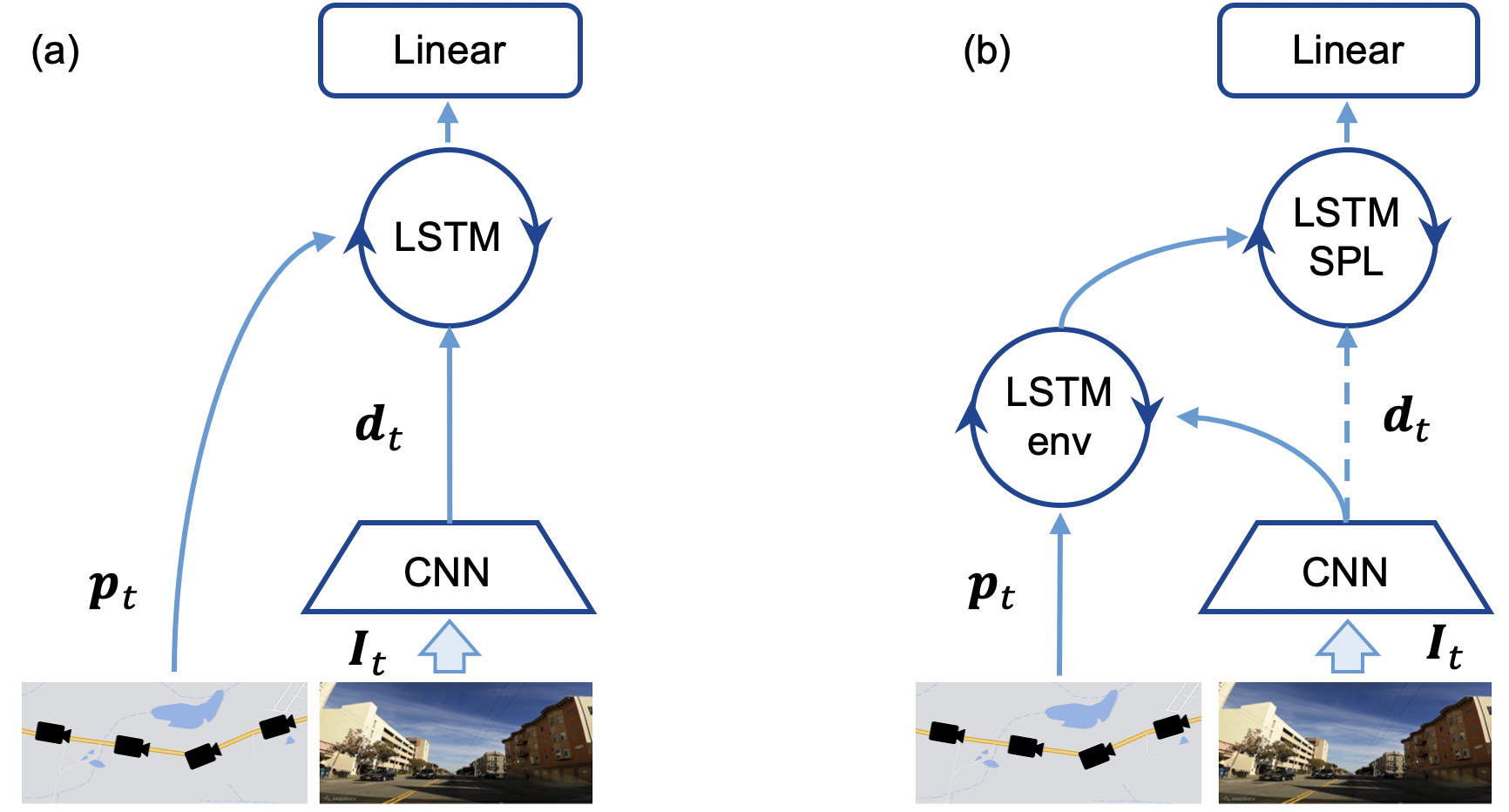}
  \caption{(a) Baseline network. (b) Proposed CNN+LSTM architecture for positional $\mathbf{p}_t$ and visual $\mathbf{d}_t$ sequential place learning (SPL).}
  \vspace{-0.4cm}
  \label{fig:arch}
\end{figure}

\subsection{Neural Network Architecture}
\label{subsec:arch}

The overall CNN+LSTM architecture for \textit{global image description} and \textit{sequential place learning} is shown in Fig. \ref{fig:arch}(b) comprising a single CNN vision module and 2 recurrent LSTM cells, along with a baseline network (a) with an LSTM.

\textbf{Global place description:} Given an image sequence $I_t$ of an environment, we apply a CNN function on each input image to obtain compact $n$-dimensional \textit{global image descriptors} $\mathbf{d}_t$, where $n$ $\in$ $\mathbb{N}^+$ is a function of the CNN model. These representations can be learned through conventional training via backpropagation. After training, the CNN will have visually encoded the entire environment within its network weights. 

\textbf{Sequence place learning (SPL):} Depending on the type of CNN chosen, and its particular training requirements, the overall CNN+LSTM training stage can be alleviated by using a pre-trained CNN function such as NetVLAD \citep{netvlad}, heavily used in visual place recognition research for extracting robust global image descriptors. Thus, the CCN module can be frozen while the LSTM cell are trained via BPTT \citep{WERBOS1988339}. In Fig. \ref{fig:arch}(b), the environment-specific (env) LSTM cell receives the CNN visual features, temporally concatenated with corresponding positional encodings. This allows our model to capture locale-specific visual and topology features that can then be used to feed a LSTM cell for SPL. The (SPL) LSTM cell receives a direct (skip) connection from the CNN module output, concatenated with the hidden states of the environment-specific LSTM output. Finally, a single linear layer (or multi-layer perceptron (MLP)) receives the hidden states from the (SPL) LSTM cell for semi-supervised training (see Fig. \ref{fig:arch}).

\textbf{Implementation details:} We experimented with several (pre-trained and trained from scratch) CNN vision models. Not surprisingly, we found that NetVLAD \citep{netvlad}, a VGG-16 \citep{vgg} based architecture, provided better results on average than other vanilla models including AlexNet \citep{krizhevsky2014one}, VGG-16, ResNet-18 \citep{he2016deep}, SqueezeNet \citep{iandola2016squeezenet}, DenseNet-161 \citep{Huang_2017_CVPR}. Thus, we used the best model (NetVLAD+whitening, trained on Pittsburgh 30k \citep{7054472}), with feature dimension of $n=4096$, from the official MATLAB implementation\footnote{https://github.com/relja/netvlad} for all state-of-the-art comparisons in Section \ref{sec:results}. The NetVLAD image descriptors $\mathbf{d}_t$ were obtained using 224$\times$224 RGB image observations, and the 2D positional information $\textbf{p}_t$ was encoded using a 2-\textit{d} vector. Depending on the type of motion data provided by each dataset, we could use any source of motion estimation including GPS, odometry or SfM. For all experiments, we standardize $\textbf{p}_t$ ($\mu=0$, $\sigma=1$) prior to feeding the network. For the recurrent networks, two vanilla single-cell LSTMs, with 512 units each, were trained end-to-end. The number of units $N$ of the linear output layer was set to be equal to the total number of frames of the dataset minus TW length.


\begin{figure*}[!t]
  \centering
  \includegraphics[width=0.32\linewidth]{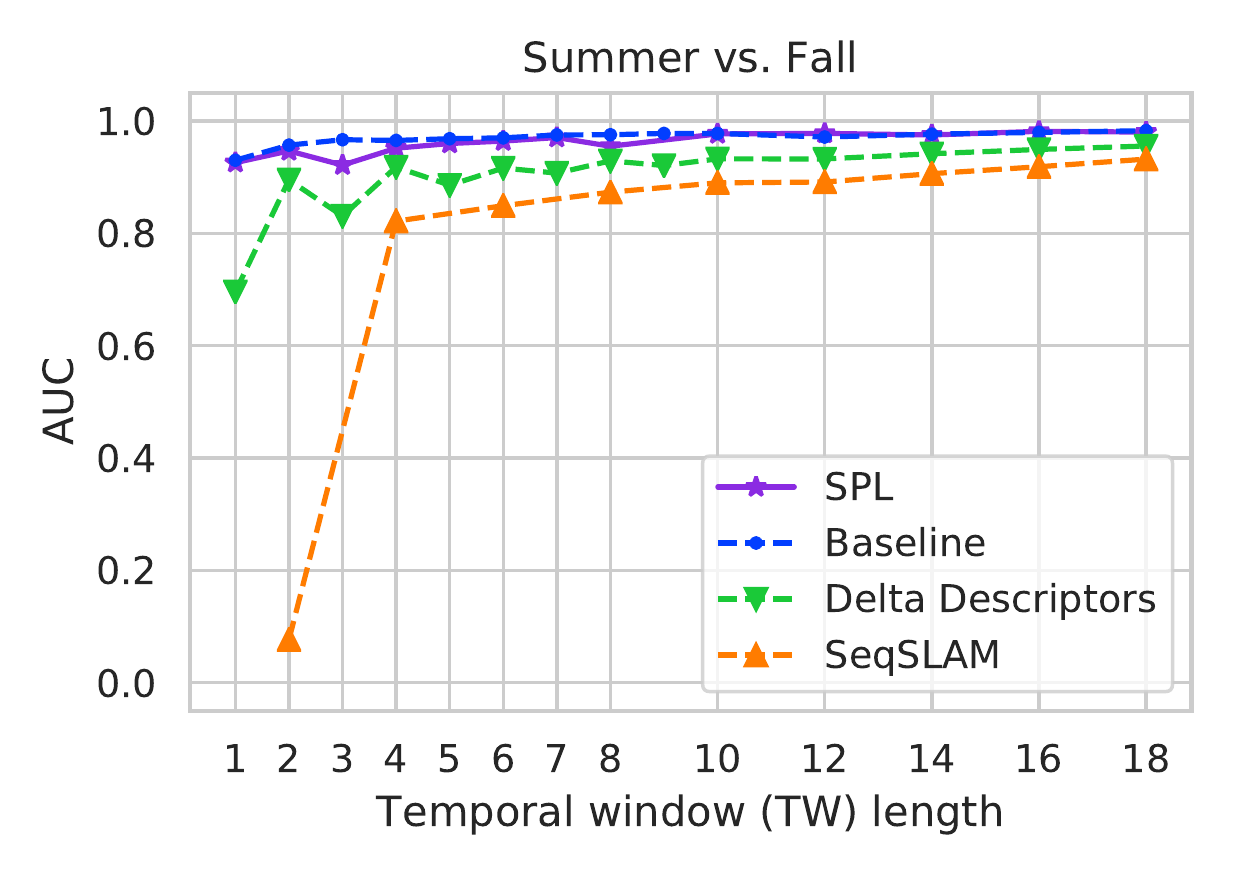}
  \includegraphics[width=0.32\linewidth]{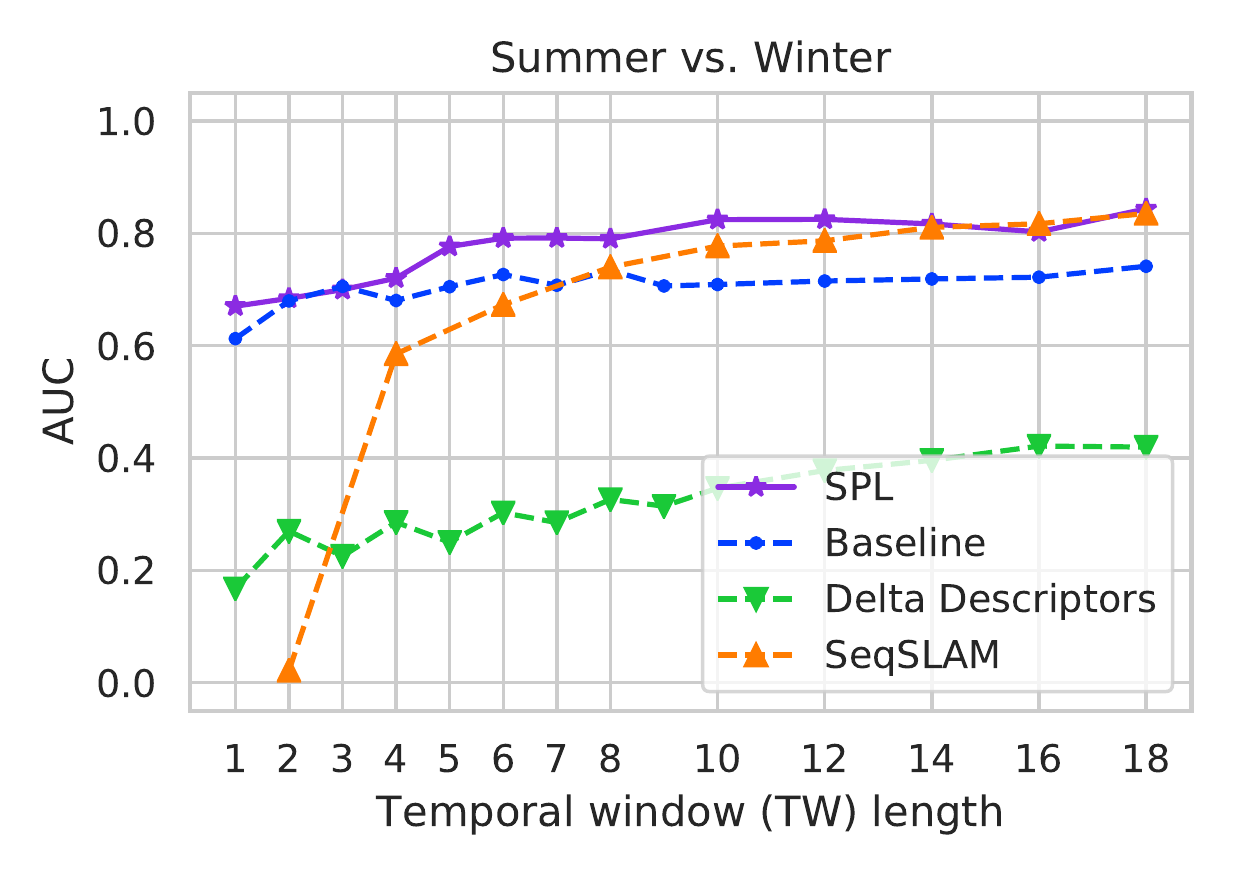}
  \includegraphics[width=0.32\linewidth]{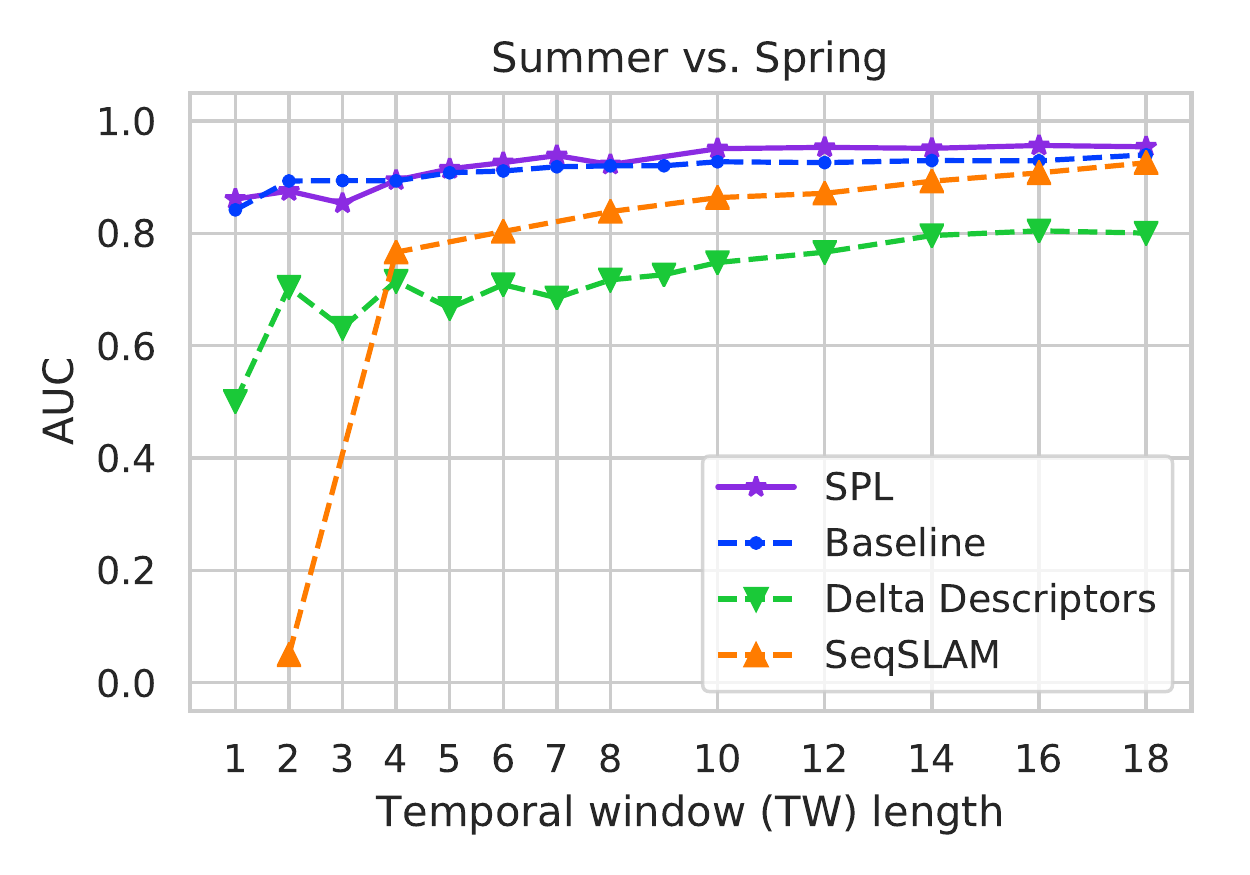}
  \caption{\textbf{Influence of temporal window (TW)} on multi-frame methods and generalization from \textit{summer} to \textit{fall}, \textit{winter} and \textit{spring} conditions of Nordland.}
  \label{influence}
\end{figure*}

\begin{figure*}[!t]
  \centering
  \includegraphics[width=0.32\linewidth]{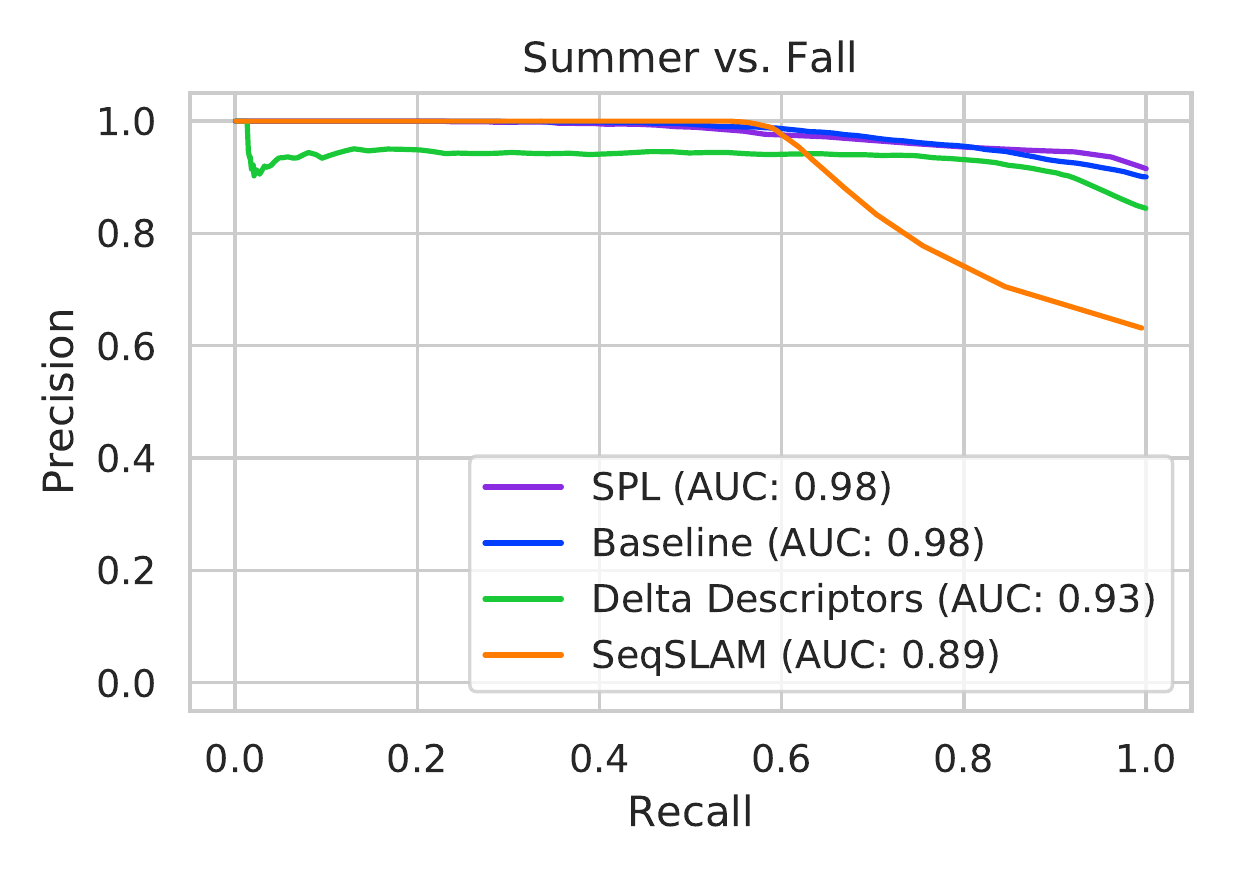}
  \includegraphics[width=0.32\linewidth]{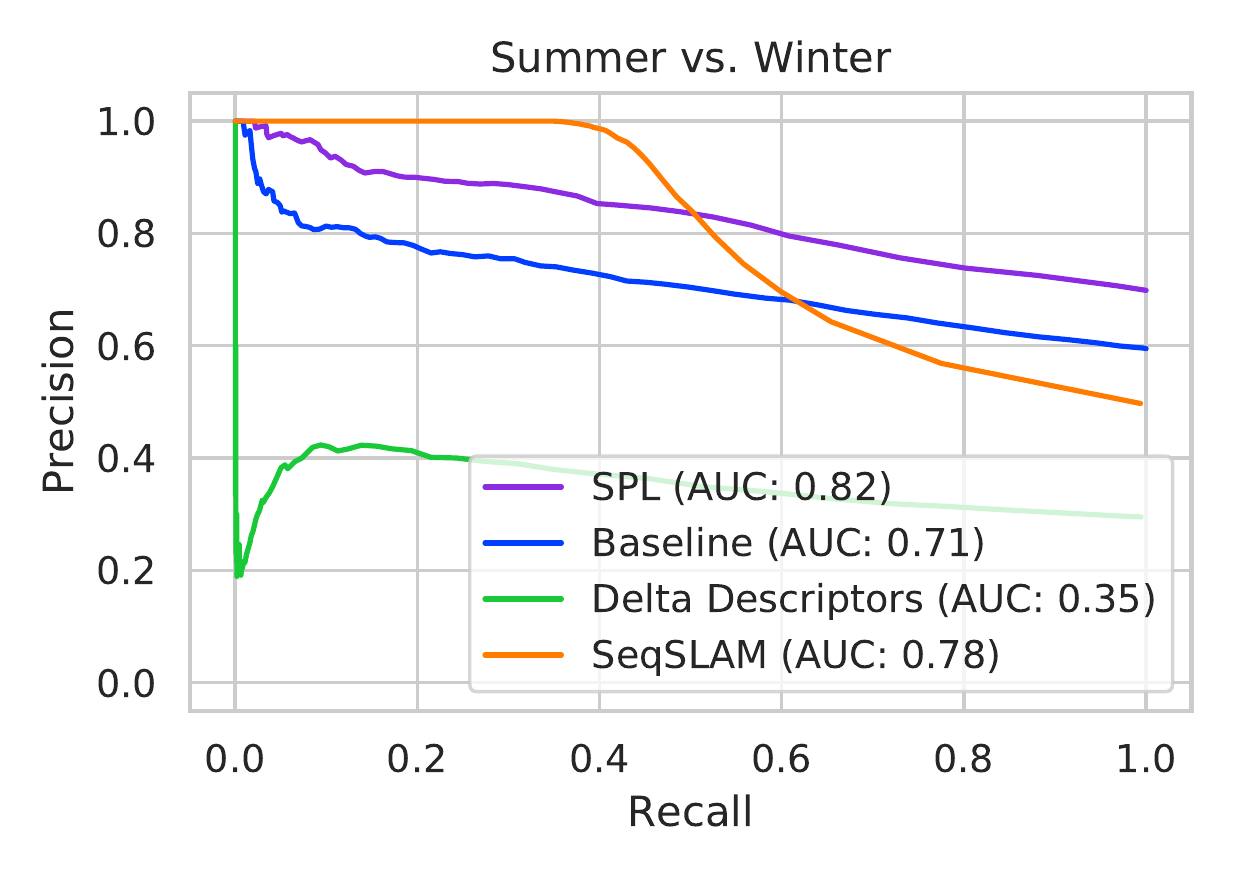}
  \includegraphics[width=0.32\linewidth]{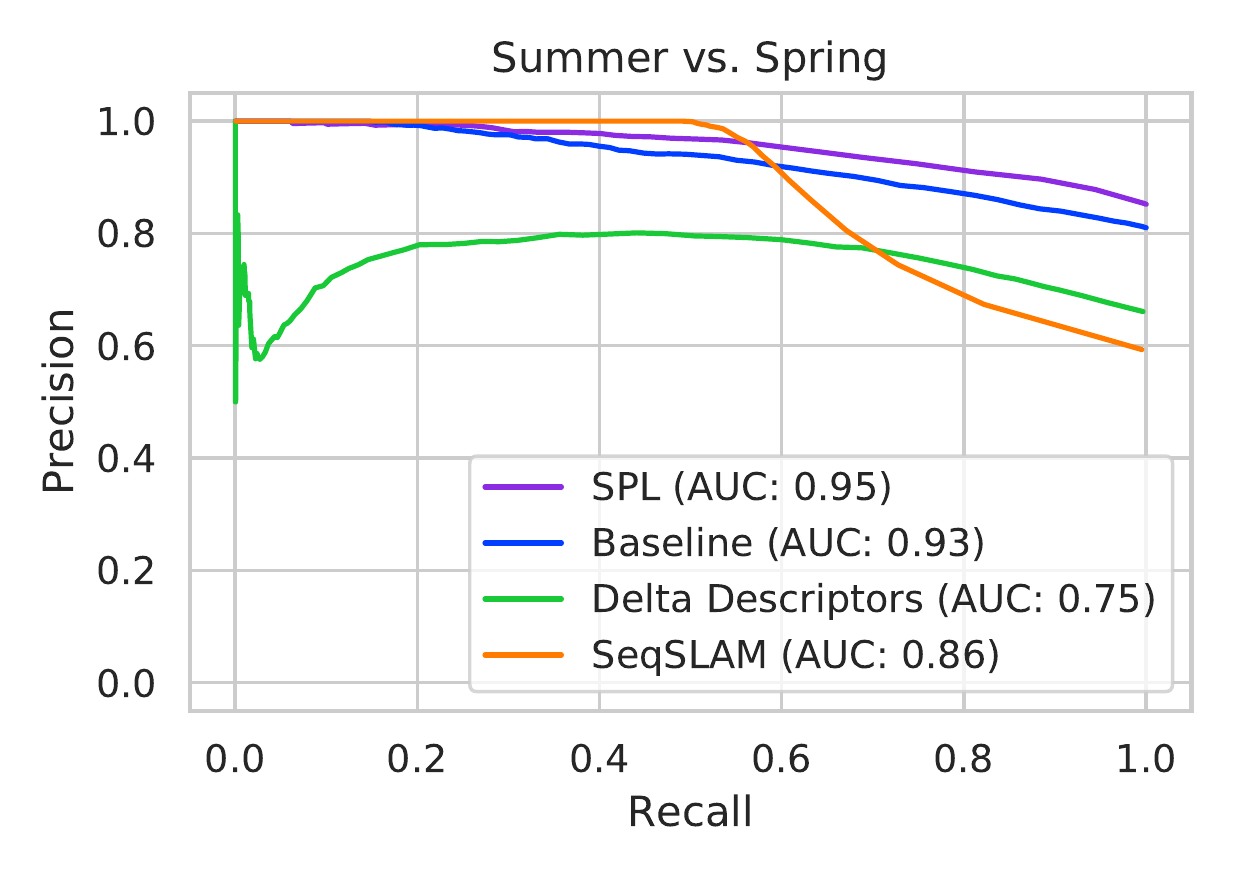}
  \caption{\textbf{Precision-recall curves and AUC metrics on the Nordland dataset} using a localization radius of 10 frames and TW of 10.}
  \label{n_pr10}
\end{figure*}

\textbf{Learning hyperparameters:} The cost function of the sequential inference task was sent to the Adam learning gradient algorithm \citep{adam} with plateau scheduler for reducing the learning rate (lr), with $\texttt{weight\_decay}=0$, $\texttt{initial\_lr}=0.001$, and $\texttt{min\_lr}=10^{-6}$. The weight assigned to the positional encodings was 500, found through cross-validation. The number of training epochs used vary depending on the dataset size. In this paper, we trained our models using 200 up to 3,000 epochs per environment, comprising between 200 and 35K+ consecutive frames, respectively, with batch sizes equal to the number of frames for stable training. After training, our models were capable of temporally integrating both vision and positional data, even using short TW lengths. 

\subsection{Sequence Processing and Temporal Window (TW)}
\label{subsec:tw}

The LSTM cell in Fig. \ref{fig:arch} learn a function to perform sequence inference based on image and positional sequential data. To enable convergence during training, we transform these sequences of observations and states into multiple examples using a sequential processing strategy over a TW length. Given a sequence with $N$ time steps $I_t=[I_1, I_2,..., I_t,..., I_N]$ (representing a sequence of raw images or positional encodings) we split this into multiple samples (sub-sequences) of TW consecutive time steps, resulting in the following $N$-$TW$ samples: $S_1=[I_1, I_2,...,I_{TW}]$, $S_2=[I_2, I_3,...,I_{TW+1}]$, ...
$S_{N-TW}=[I_{N-TW}, I_{N-TW+1},...,I_{N-1}]$. We feed our model with a \texttt{batch\_size} number of these temporally synchronized, consecutive (image and positional) samples, and highlight that our implementation is robust to training and deployment with shuffled samples, and also to velocity inconsistency between training and query traversals. We provide experiments on asynchronous datasets to demonstrate this capability.



\section{Experimental Setup}
\label{sec:setup}

\subsection{Sequence-based Datasets}

\textbf{Nordlandsbanen:} The minute by minute, season by season Nordland Railway dataset\footnote{https://nrkbeta.no/2013/01/15/} was recorded over a 729 km train journey in Norway, providing four huge $\sim$10-hour video streams, one for every season, at 25 FPS and 1920$\times$1080 resolution. Each video file is synchronized in order to make the train appear to be at the same place at the same time. It also provides synchronized GPS data for each traversal. On this dataset, we conducted two main experiments by extracting individual frames out of videos at 1 FPS and 0.1 FPS, resulting in 35768 and 3577 frames, respectively. We trained our model on \textit{summer} conditions and tested on the remaining (\textit{fall}, \textit{winter}, \textit{spring}), and corresponding GPS data was encoded to represent positional information as discussed in Section \ref{subsec:arch}.

\textbf{Oxford RobotCar:} This dataset \citep{robotcar} was collected on a car platform traversing over 100+ times over a 10 km route in Oxford, UK, over a 1-year period, capturing diverse weather, season, and dynamic urban conditions. We selected 1000 temporally synchronized frames out of three traversals\footnote{2014-12-09-13-21-02, 2015-05-19-14-06-38, 2014-12-10-18-10-50 in \citep{robotcar}}, referred here as \textit{overcast} for training, and \textit{sunlight} and \textit{night} for testing, with corresponding GPS data. On this particular dataset, we used 1280$\times$960 RGB images from the center side of the Bumblebee XB3 trinocular camera but without applying undistortion, thus, slightly increasing the difficulty for recognizing places.

\textbf{Gardens Point:} The day and night with lateral pose change (right, left) Gardens Point Walking dataset \citep{gardens} consists of three synchronized traversals with 200 images at 960$\times$540 resolution each. We conduct several experiments for viewpoint and appearance changes on this dataset by training and testing on different traversal combinations as detailed in Section \ref{sec:results}. This dataset does not provide motion information but we use synthetic time-series data provided in \citep{chancan2020deepseqslam}.

\textbf{St Lucia:} The St Lucia Multiple Times of Day dataset \citep{Glover2010ICRA} was collected with a forward facing webcam, attached to the roof of a car, through the suburb of St Lucia, Queensland, Australia. The route was traversed ten times during multiple days to capture the difference in appearance between early morning and late afternoon. GPS data is included for each trip and synchronized with 640$\times$480 RBG images at 15 FPS. On this dataset, we demonstrate that our approach is robust to strong velocity inconsistencies of the car between training and query traversals. We use the first 4,000 frames of each traversal, covering the full $\sim$10-km perimeter of the suburb.



\subsection{Precision-Recall, Average-Precision and Tolerances}

\textbf{Precision-Recall and Average-Precision:} On deployment, likelihood scores for each query image w.r.t. the reference traversal are produced by the linear output layer of our model. These values were then used to compute corresponding precision-recall (PR) curves, which can then be used to calculate other types of metrics such as area under the curve (AUC). We report both PR curves and AUC metrics for all experiments and datasets.

\textbf{Coarse Localization Radius Tolerance:} Except for St Lucia, we report results on a range of ground-truth error tolerance from 1 up to 50 frames away from the correct match for performance analysis in Section \ref{sec:results}. On St Lucia, we consider a localization radius tolerance of 20 meters.

\section{Results}
\label{sec:results}

Here we demonstrate that our approach is capable of learning sequence inference from a single traversal of a route, while accurately generalizing to multiple traversals of that route under very different visual conditions. Also, on the most challenging asynchronous dataset, St Lucia, with velocity inconsistencies, our approach gets 96.65\% AUC while SeqSLAM and Delta Descriptors struggle with 27.39\% AUC and 57.47\% AUC, respectively.

\subsection{Influence of Temporal Window (TW)}

\begin{figure*}[!t]
  \centering
  \includegraphics[width=0.32\linewidth]{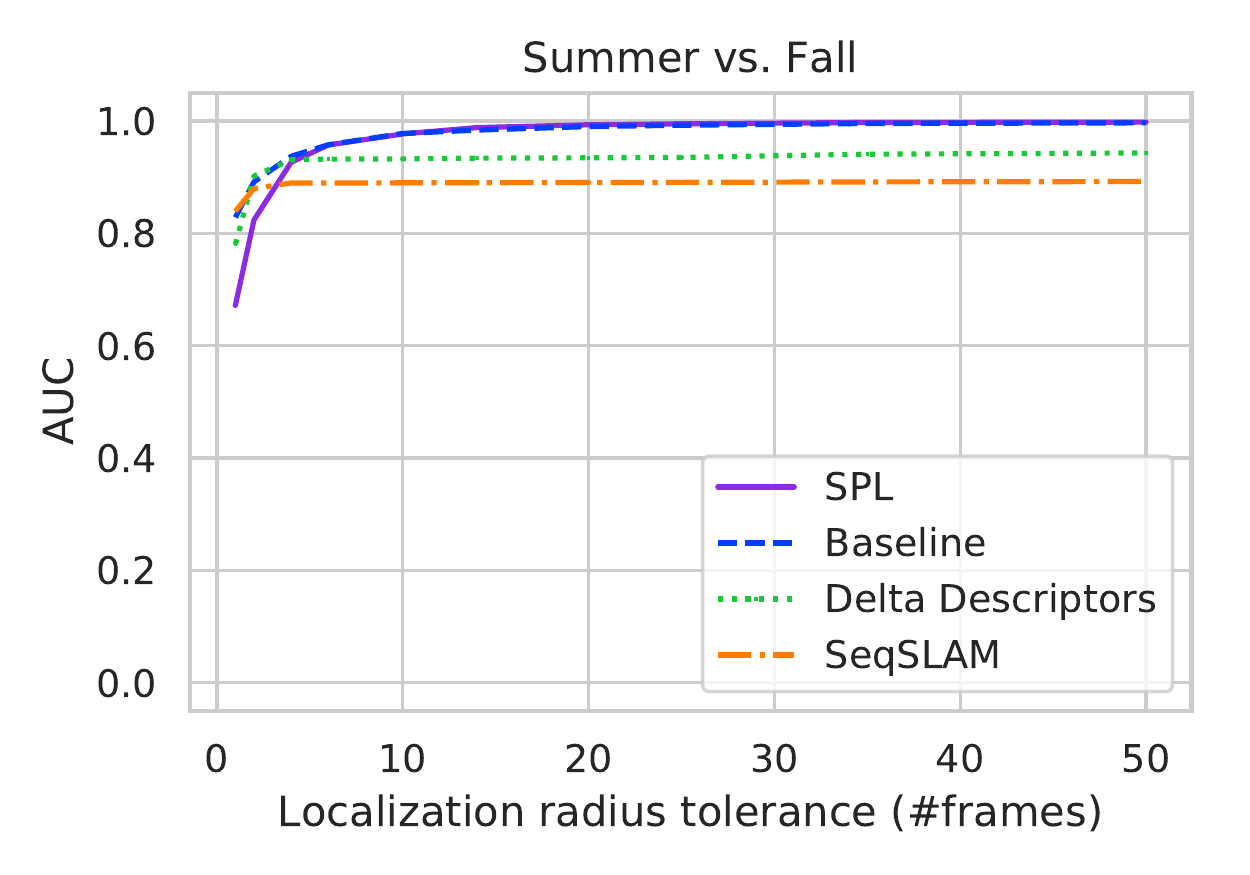}
  \includegraphics[width=0.32\linewidth]{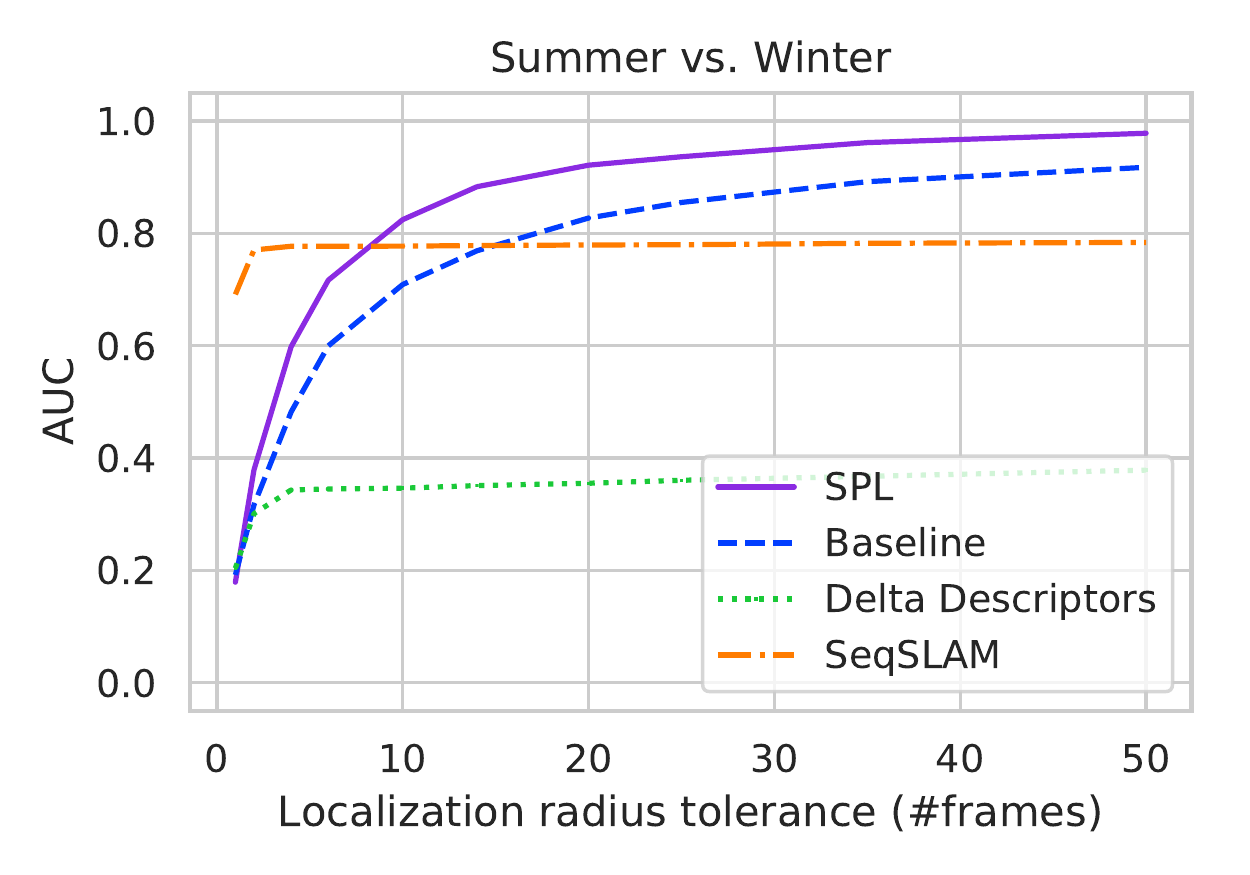}
  \includegraphics[width=0.32\linewidth]{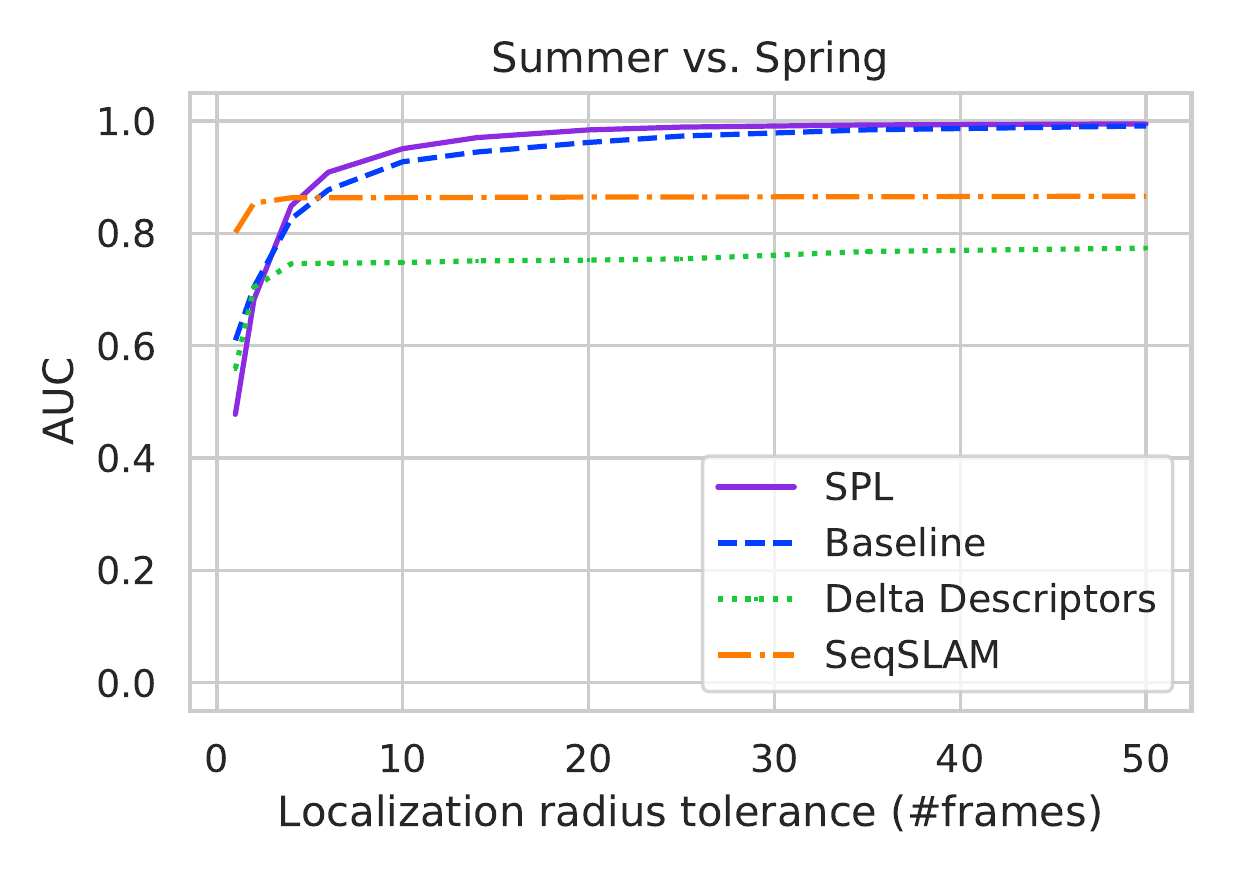}
  \caption{\textbf{AUC metrics vs. localization radius tolerance on the Nordland dataset} with TW of 10.}
  \label{n_tol}
\end{figure*}

\begin{table*}
\caption{Space and Time Synchronized Datasets: AUC (\%) Results for Localization Radius of $2/10/50$ and TW of 10}
\begin{center}
\label{gno_tol}
\begin{tabular}{|c||c|c||c|c||c|c|}
\hline
\multirow{2}{*}{\textbf{Method}} & \multicolumn{2}{c||}{\textbf{Gardens Point}} & \multicolumn{2}{c||}{\textbf{Nordlandsbanen}} & \multicolumn{2}{c|}{\textbf{Oxford RobotCar}} \\
\cline{2-7}
& Day (L)-Night (R) & Day (R)-Day (L) & Summer-Fall & Summer-Winter & Overcast-Sunlight  & Overcast-Night \\
\hline\hline
SeqSLAM \citep{seqslam} & 5.2 / 12.8 / 46.2 & 43.2 / 56.5 / 65.8 & 88.0 / 89.0 / 89.2 & \textbf{77.0} / 77.7 / 78.3 & 9.4 / 16.9 / 18.6 & 1.3 / 3.9 / 12.0 \\
\hline
Delta D. (DD) \citep{deltad} & 79.8 / 93.8 / 96.6  & 96.1 / 99.5 / 99.5 & \textbf{90.2} / 93.2 / 94.3 & 30.0 / 34.6 / 37.8 & 4.9 / 16.1 / 27.2 & \textbf{87.1} / 87.1 / 87.1 \\
\hline
Baseline (BL) \citep{chancan2020deepseqslam} & \textbf{100} / \textbf{100} / \textbf{100}  & \textbf{100} / \textbf{100} / \textbf{100} & 89.1 / \textbf{97.8} / 99.6 & 31.6 / 70.9 / 91.7 & \textbf{78.4} / 96.7 / \textbf{100} & 69.6 / 93.1 / 99.3 \\
\hline
\textbf{SPL} & \textbf{100} / \textbf{100} / \textbf{100}  & \textbf{100} / \textbf{100} / \textbf{100} & 82.3 / \textbf{97.8} / \textbf{99.8} & 37.8 / \textbf{82.4} / \textbf{97.8} & 66.9 / \textbf{97.9} / \textbf{100} & 53.2 / \textbf{96.0} / \textbf{99.9} \\
\hline
\end{tabular}
\end{center}
\end{table*}

\begin{table*}
\caption{Asynchronous Datasets: AUC (\%) Results for Localization Radius of 20 meters and TW of 10}
\begin{center}
\label{st_auc}
\begin{tabular}{|c||c|c|c|c|c|c|c|c|}
\hline
\multirow{2}{*}{\textbf{Method}} & \multicolumn{8}{c|}{\textbf{St Lucia}} \\
\cline{2-9}
& 100909\_1000 & 100909\_1210 & 100909\_1410 & 110909\_1545 & 180809\_1545 & 190809\_1410 & 210809\_1000 & 210809\_1210 \\
\hline\hline
SeqSLAM \citep{seqslam} & 61.64 & 50.04 & 48.74 & 27.39 & 44.12 & 61.32 & 44.51 & 38.19 \\
\hline
DD \citep{deltad} & 97.89 & 92.71 & 78.03 & 57.47 & 74.58 & 67.28 & 98.98 & 92.79 \\
\hline
\textbf{SPL} & \textbf{99.94} & \textbf{99.68} & \textbf{99.54} & \textbf{96.65} & \textbf{98.99} & \textbf{99.61} & \textbf{99.78} & \textbf{97.07} \\
\hline
\end{tabular}
\end{center}
\end{table*}

In Fig. \ref{influence} we demonstrate how classical methods such as SeqSLAM \citep{seqslam} and Delta Descriptors \citep{deltad} struggle to produce accurate results when using short TW lengths, especially under strong visual changes as discussed throughout the paper, while learning-based methods achieve competent AUC results above 60\%. We used a localization radius tolerance of 10 frames for producing this figure at every single value of TW, which is lower than in previous research \citep{chancan2020deepseqslam}. For SeqSLAM, we use the official MATLAB implementation\footnote{\href{https://github.com/OpenSLAM-org/openslam_openseqslam}{https://github.com/OpenSLAM-org/openslam\_openseqslam}} with default parameters (only changing TW or $d_s$ as needed); SeqSLAM only works using an even number for $d_s$ though. For Delta Descriptors, we use the Python code provided by the authors in \citep{deltad} with default parameters also varying TW only as needed. We highlight that the baseline \citep{chancan2020deepseqslam}, is getting competitive performance with its generalized method but under \textit{summer}-\textit{winter} changes its performance remains constant at 70\% AUC, even using larger TWs, while our approach and SeqSLAM achieve over 80\% AUC. However, when using a TW of 2 frames (Fig. \ref{influence}-Middle), SeqSLAM and Delta Descriptors get 2\% and 27\% AUC, respectively, while ours gets near 70\%. 

\subsection{Comparisons on Space and Time Synchronized Datasets}
\label{subsec:sota}

In Fig. \ref{n_pr10} we report the PR curves for a TW of 10 obtained from the results on Fig. \ref{influence}, with corresponding AUC metrics, considering a localization radius tolerance of 10 frames. On the most challenging \textit{summer}-\textit{winter} conditions changes, we note that SeqSLAM performs competitively at this TW with 78\% AUC on \textit{winter} (Fig. \ref{n_pr10}-middle), but our model outperforms it with 82\% AUC overall. At the same conditions, Delta Descriptors is shown to perform poorly due to both image resolution and huge sampling rates sensitivity on this particular dataset settings. We will show, however, that its performance gets better on the other dataset configurations for fair comparisons, but we wanted to highlight this particular limitation among classical methods. Nevertheless, on the other less challenging conditions changes (\textit{fall}, \textit{spring}), we show all these models achieve competent results approaches 75\%.

More generally, in Fig. \ref{n_tol}, we show how the AUC spectrum changes against localization radius tolerances between 1 to 50 frames, while all models are using a standard value of TW=10; typically used on place recognition research. We also highlight that TW of 10 effectively captures the overall maximum AUC results of each method, as shown in Fig. \ref{influence}. In Table \ref{gno_tol} we summarize the main AUC results corresponding to our three datasets for a TW of 10 and localization radius of $2/10/50$ frames. Our model consistently attains higher AUC values than classical methods at challenging visual transitions such as \textit{day}-\textit{night} and \textit{summer}-\textit{winter}. In the Appendix we report all the PR curves from where the AUC metrics presented in Table \ref{gno_tol} were obtained, along with the training curves of our approach.

\begin{table*}
\begin{center}
\caption{Same Front-End for all Methods: AUC Results using NetVLAD on the Gardens Point dataset}
\begin{tabular}{|c|c||c|c|c|c|c|c|c|c|c|c|c|}
\hline
\multicolumn{2}{|c||}{\textbf{Traversal}} & \multicolumn{11}{c|}{\textbf{Methods with NetVLAD Front-End}} \\
\hline
Ref & Query & Pairwise & ABLE \citep{arroyo2015towards} & ISM \citep{vysotska2015efficient} & OPR \citep{vysotska2015lazy} & VPR \citep{vysotska2017relocalization} & HMM \citep{hansen2014visual} & SeqSLAM & MCN \citep{8756053} & DD & BL & \textbf{SPL} \\
\hline\hline
D-L & N-R & 0.41 & 0.79 & 0.61& 0.48 & 0.29 & 0.02 & 0.15 & 0.43 & 0.93 &  0.99 & \textbf{1} \\
\hline
D-R & D-L  & 0.98 & \textbf{1} & 0.69 & 0.69 & 0.69 & 0.33 & 0.68 & 0.99 & 0.99 &  0.99 & \textbf{1} \\
\hline
D-R & N-R  & 0.52 & 0.8 & 0.64 & 0.64 & 0.47 & 0.20 & 0.30 & 0.54 & 0.98 &  0.99 & \textbf{1} \\
\hline
\end{tabular}
\label{tab:nvlad}
\end{center}
\end{table*}

\begin{table*}
\caption{Influence of Viewpoint and Appearance Changes: Maximum Recall (\%) at 100\% Precision}
\begin{center}
\begin{tabular}{|c||c|c||c|c||c|c|}
\hline
\textbf{Changes} & \multicolumn{2}{c||}{\textbf{Viewpoint}} & \multicolumn{2}{c||}{\textbf{Appearance}} & \multicolumn{2}{c|}{\textbf{Viewpoint \& Appearance}} \\
\hline
\textbf{Dataset} & \textbf{Gardens Point} & \textbf{CSU-1} & \textbf{Gardens Point} & \textbf{Nordland} & \textbf{Gardens Point} & \textbf{CSU-2} \\
\hline
Reference-Query & Day (L)-Day (R) & Day (L)-Day (R) & Day (R)-Night (R)  & Summer-Winter & Day (L)-Night (R)  & Day (L)-Day (R) \\
\hline\hline
FAB-MAP \citep{cummins2008fab} & 2.0 & 14.3 & - & - & - & - \\
\hline
VLAD-based \citep{lowry2018lightweight} & 19.5 & 59.0 & 2.5 & 2.0 & - & 11.0 \\
\hline
SeqSLAM \citep{seqslam} & 1.0 & 25.9 & 3.0 & 4.6 & - & 16.0 \\
\hline
SMART \citep{smart} & 13.0 & 12.5 & 5.0 & 4.4 & - & 1.0 \\
\hline
CNN+Seq \citep{chen2014convolutional} & 45.0  & 67.9 & 48.0 & 9.0 & 14.0 & 41.0 \\
\hline
RISF \citep{avp} & 46.0  & 91.0 & 63.0 & \textbf{22.9} & 67.5 & 90.0 \\
\hline
Baseline \citep{chancan2020deepseqslam} & 99.0  & \textbf{100} & 99.0 & 1.2 & 99.0 & \textbf{100} \\
\hline
\textbf{SPL} & \textbf{100}  & \textbf{100} & \textbf{100} & 2.5 & \textbf{100} &\textbf{100} \\
\hline
\end{tabular}
\label{tab:changes}
\end{center}
\end{table*}

\subsection{Comparisons on Asynchronous Datasets}

In Table \ref{st_auc}, we report the results on the St Lucia dataset, recorded under variable velocities of the vehicle as discussed in Sections \ref{sec:setup} and \ref{subsec:tw}. We show AUC values for late morning and all afternoon times (totaling 8 query traversals), given that our reference traversal was recorded early morning around 08:45 a.m. ($190809\_0845$). In the Appendix, however, we show the full PR curves and qualitative results on the full 10 subsets for each method. Our approach again significantly outperforms other methods on all the subsets.

\subsection{Comparison to Ten Methods using the Same Front-End}

All the results reported in Section \ref{subsec:sota} were obtained using their original fron-end methods for global image description. SeqSLAM used classical sum of absolute differences (SAD), and the others (Delta Descriptors, Baseline, and our approach) by default use the best NetVLAD model as we described in Section \ref{subsec:arch}. In Table \ref{tab:nvlad}, we report additional supporting comparisons against single-frame vanilla NetVLAD (pairwise) and other 6 multi-frame filtering methods (ABLE \citep{arroyo2015towards}, ISM \citep{vysotska2015efficient}, OPR \citep{vysotska2015lazy}, VPR \citep{vysotska2017relocalization}, HMM \citep{hansen2014visual}, SeqSLAM \citep{seqslam}, MCN \citep{8756053}) that received the pairwise similarity matrix, obtained from NetVLAD \citep{netvlad}, for sequence filtering according to \citep{8756053}. In addition to SeqSLAM, Delta Descriptors (DD) and Baseline (BL), in Table \ref{tab:nvlad} we present the full comparison all these ten methods on the Gardens Point dataset. AUC results are calculated with the same localization tolerance used in \citep{8756053}. Our model outperforms all the others with 100\% recall at 100\% precision on all the required reference-query combinations of subsets: day-left (D-L), day-right (D-R) and night-right (N-R).

\subsection{Robustness to Viewpoint and Appearance Changes}

In Table \ref{tab:changes}, we reproduce the benchmark presented in \citep{avp} where the authors compare their model under changes in viewpoint, appearance and both types of changes at the same time on three datasets. This allows us to compare our approach against five other methods such as FAB-MAP \citep{cummins2008fab}, VLAD-based \citep{lowry2018lightweight}, SMART \citep{smart}, CNN+Seq \citep{chen2014convolutional}, Robust Image-sequence-based framework (RISF) \citep{avp}, in addition to SeqSLAM and the baseline network. Our best model once again achieves precision rates of 100\% at 100\% recall on all but one condition. Where the method in \citep{avp} performs the best with 22.9\% recall, while ours gets 2.5\% recall. It is worth noting, however, that our model can achieve up to 82\% AUC on this particular condition, which on average is comparable to $\sim$ 88\% AUC of the method proposed in \citep{avp} according to their PR curves. Overall at this point we have already compared SPL against 15 place recognition systems on 5 datasets.


\begin{table}[!t]
  \caption{CPU Only: Deployment Time on the Nordland dataset}
  \label{table_time}
  \centering
  \begin{tabular}{|c||c|c|c|}
    \hline
    \textbf{Method}  & \textbf{35,768 frames} & \textbf{3,577 frames}  \\
    \hline\hline
    SeqSLAM (CPU) & 70m & 57s    \\
    \hline
    Delta Descriptors (CPU) & 51m & 33s  \\
    \hline
    \textbf{SPL} (CPU)  & \textbf{1m} & \textbf{5.7s}  \\
    \hline
  \end{tabular}
\end{table}


\subsection{High-Performance Analysis}

In Table \ref{table_time} we compare the runtime on deployment using the same CPU (Intel Core i7-8700K CPU @3.70GHz) for all these models on our two (large and medium) Nordland dataset settings, comprising 35,768 and 3,577 frames, respectively. For a medium dataset size, our CPU-based model is up to 10$\times$ faster than classical methods, while for the large dataset configuration it can be up to 70$\times$ faster. It is worth noting that these speed-ups are based on the average latency of our model running on CPU only (not on GPU). Our model SPL running on a GPU (GeForce GTX 1080Ti), shown in red in Fig. \ref{latency}, is even faster but we do not compare the latency of classical methods against our GPU deployment. In Fig. \ref{latency}, we show the performance comparison between AUC metrics and deployment latency for our medium dataset configuration, where we vary TW from 1 up to 18. All  our  models  achieve comparable results to the best SeqSLAM settings (with TW of 18), even with TW values around 5. Although we used CPU for comparison purposes, we highlight that the PyTorch \citep{NIPS2019_9015} implementation can run on CPU or GPU, contrary to classical methods that typically run on CPU.

\subsection{Qualitative Results}

\begin{figure}[!t]
  \centering
  \includegraphics[width=\linewidth]{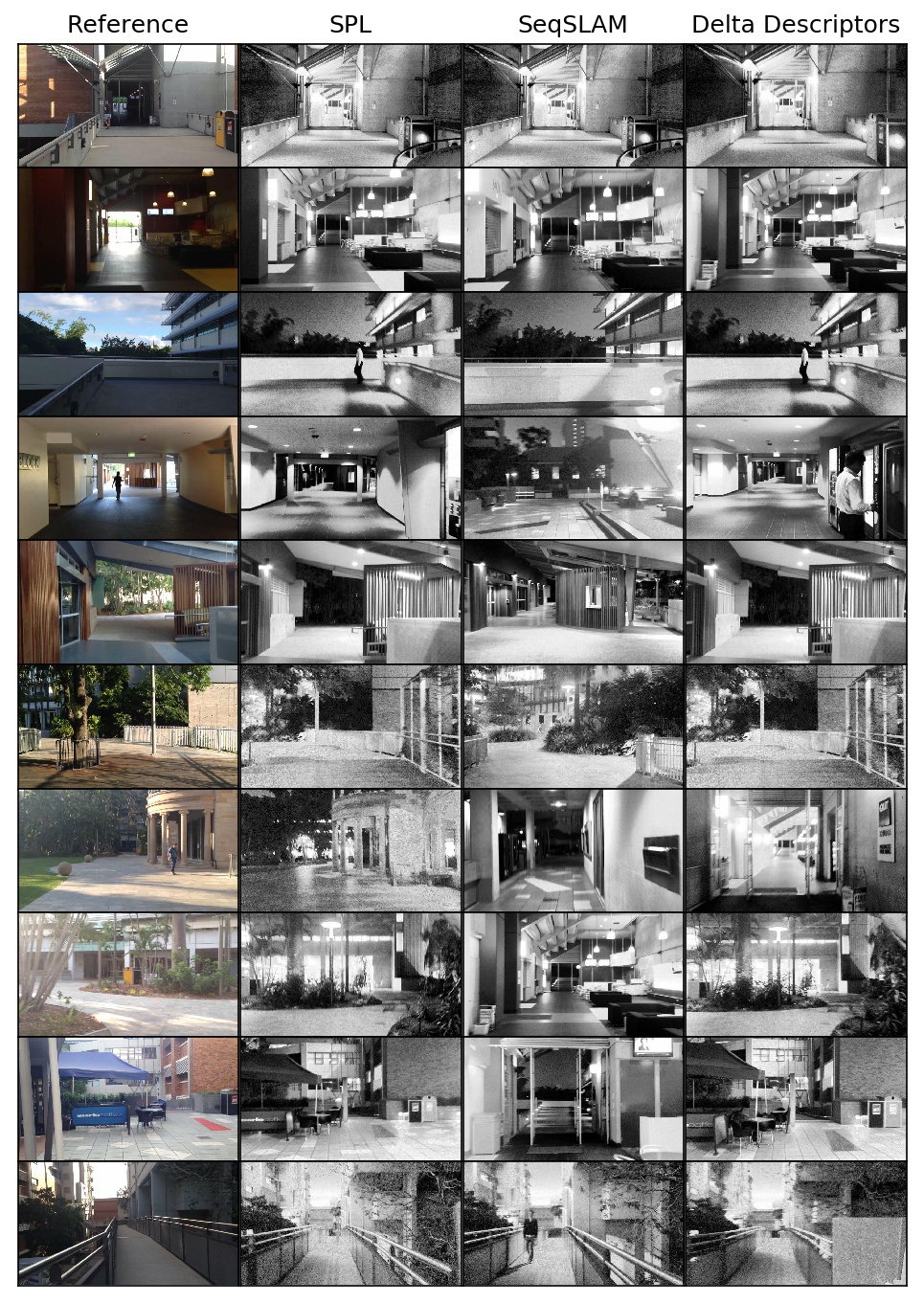}
  \caption{Deployment on Gardens Point with TW and tolerance of 10.}
  \vspace{-2mm}
  \label{gp_qualy}
\end{figure}

\begin{figure}[!t]
  \centering
  \includegraphics[width=\linewidth]{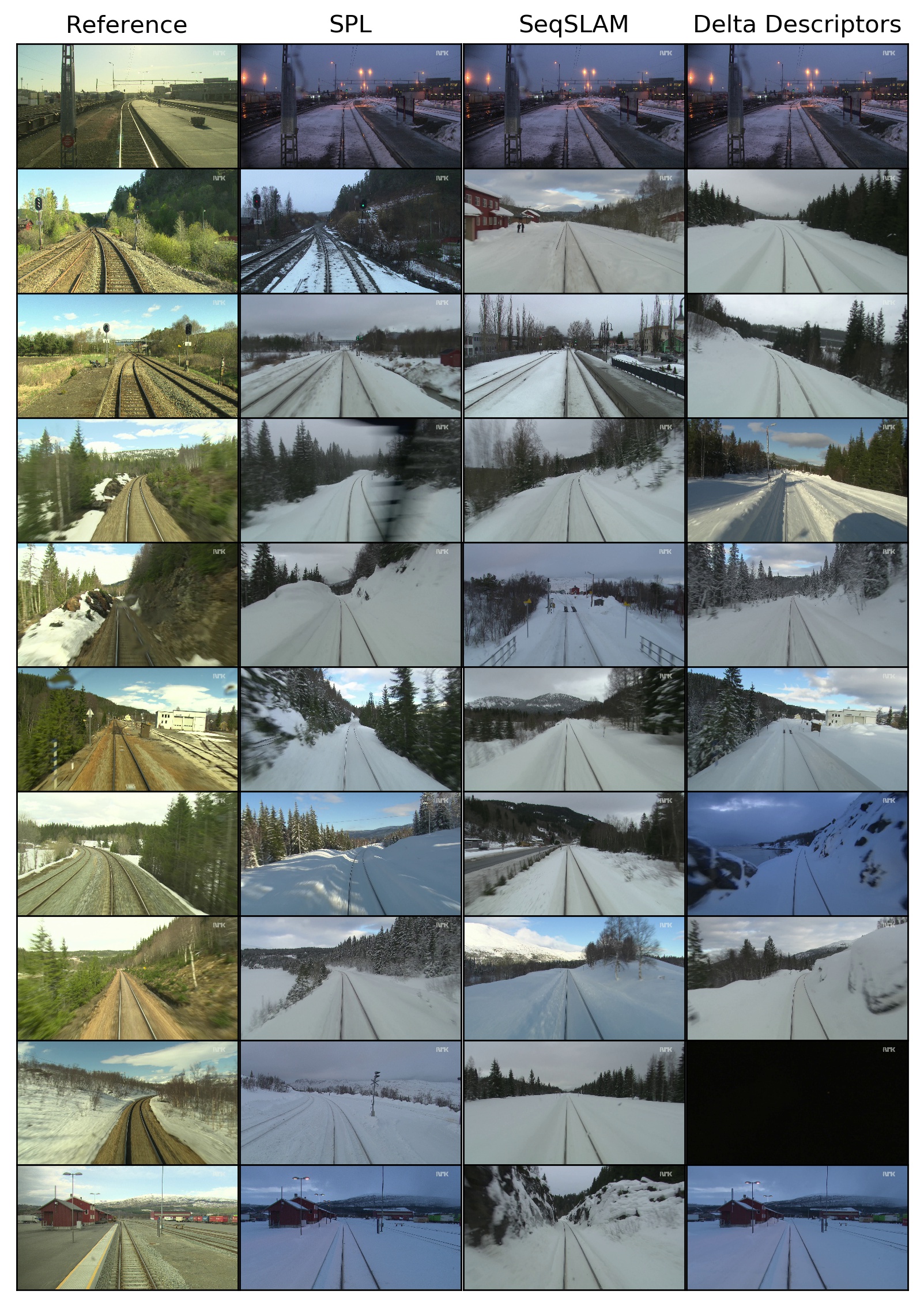}
  \caption{Deployment on Nordlandsbanen with TW and tolerance of 10.}
  \vspace{-2mm}
  \label{n_qualy}
\end{figure}

\begin{figure}[!t]
  \centering
  \includegraphics[width=\linewidth]{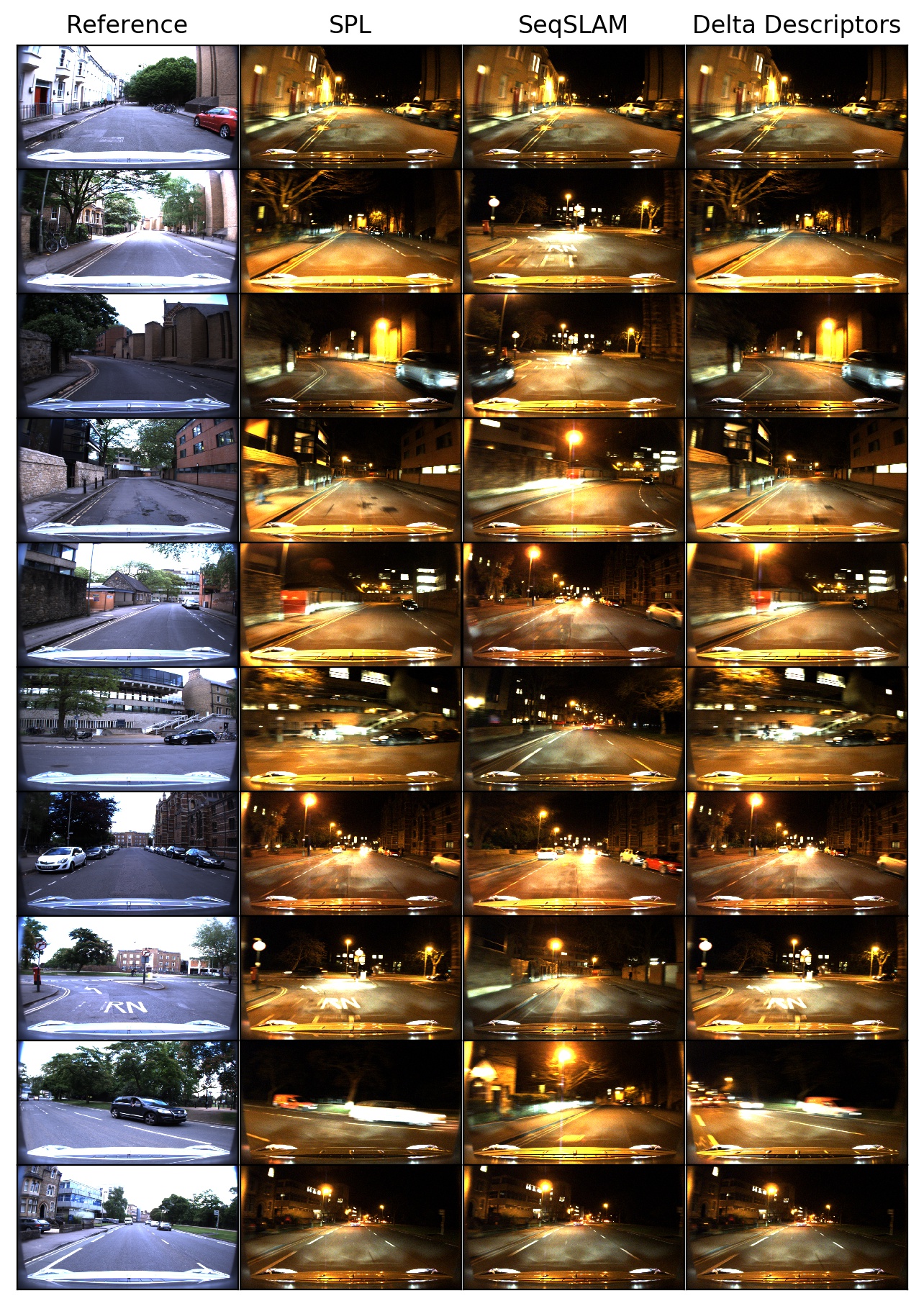}
  \caption{Deployment on Oxford RobotCar with TW and tolerance of 10.}
  \vspace{-2mm}
  \label{oxf_qualy}
\end{figure}

In Figs. \ref{gp_qualy}, \ref{n_qualy} and \ref{oxf_qualy}, we show visualizations of the raw sequence-based image matchings on the Gardens Point, Nordland and Oxford RobotCar datasets, respectively, along the entire traversal we used for each experiment. From left to right, in each figure, every column shows the top-1 match for a particular reference image (left), obtained using our approach (SPL), SeqSLAM and Delta Descriptors (right). The reference sequence was obtained by sampling 10 equally-spaced images out of the entire traversal, each traversal starting on the top and ending on the bottom of the figure. In the Appendix, we show similar qualitative results on the St Lucia dataset, but for the full 10 traversals and considering 20 equally-spaced images sampled from the entire reference traversal.





\section{Discussion and Conclusions} 
\label{sec:conclusion}


We designed and implemented an LSTM-based architecture for joint visual-and-positional encoding and sequential place learning (SPL), rather than using conventional two-stage \textit{match-then-filter} techniques. SPL was shown to be robust to extreme environmental changes and velocity inconsistency between training and query traversals, found across four large, challenging benchmark driving datasets. Our approach addressed all the main limitations of classical heuristic-based methods including high sensitivity to short temporal windows (TW) values, expensive compute and storage requirements and very limited work on learning-based systems for sequence inference using recurrent networks. It provides a strong baseline for future work in learning-based sequence filtering in the context of simultaneous localization and mapping (SLAM) and autonomous navigation research.

We proposed a generalized CNN+LSTM model that incorporates an \textit{environment-specific} LSTM cell for potentially enabling learning across different environments. This apparently simple yet more general architectural change, compared to a baseline that uses a single-cell LSTM, was found to be even more accurate and robust, while also potentially enabling further extension of our approach onto a \textit{multi-environment architecture} for training and deployment on different environments using a single model; as in related reinforcement-learning-based research for navigation \citep{withoutamap}.

Instead of relying on pre-trained CNN models, we set out to use a small two-layer CNN for exploring the end-to-end training behavior (from scratch) of our model but the results showed that the CNN component does not generalize well to drastic visual changes, which was expected since these models require a significant amount of data for effective training and generalization. We see this observation as future work to further investigate additional advantages of jointly learning visual and positional information. Finally, it is worth noting that our method has the potential for supporting the development of a full learning-based SLAM system by incorporating a geometric mapping neural network such as those in \citep{mapnet2018,bian2019depth,zhao2020towards}.






\bibliographystyle{plainnat}
\bibliography{references}



\section{Appendix}


\begin{figure*}[!h]
  \centering
  \includegraphics[width=0.36\linewidth]{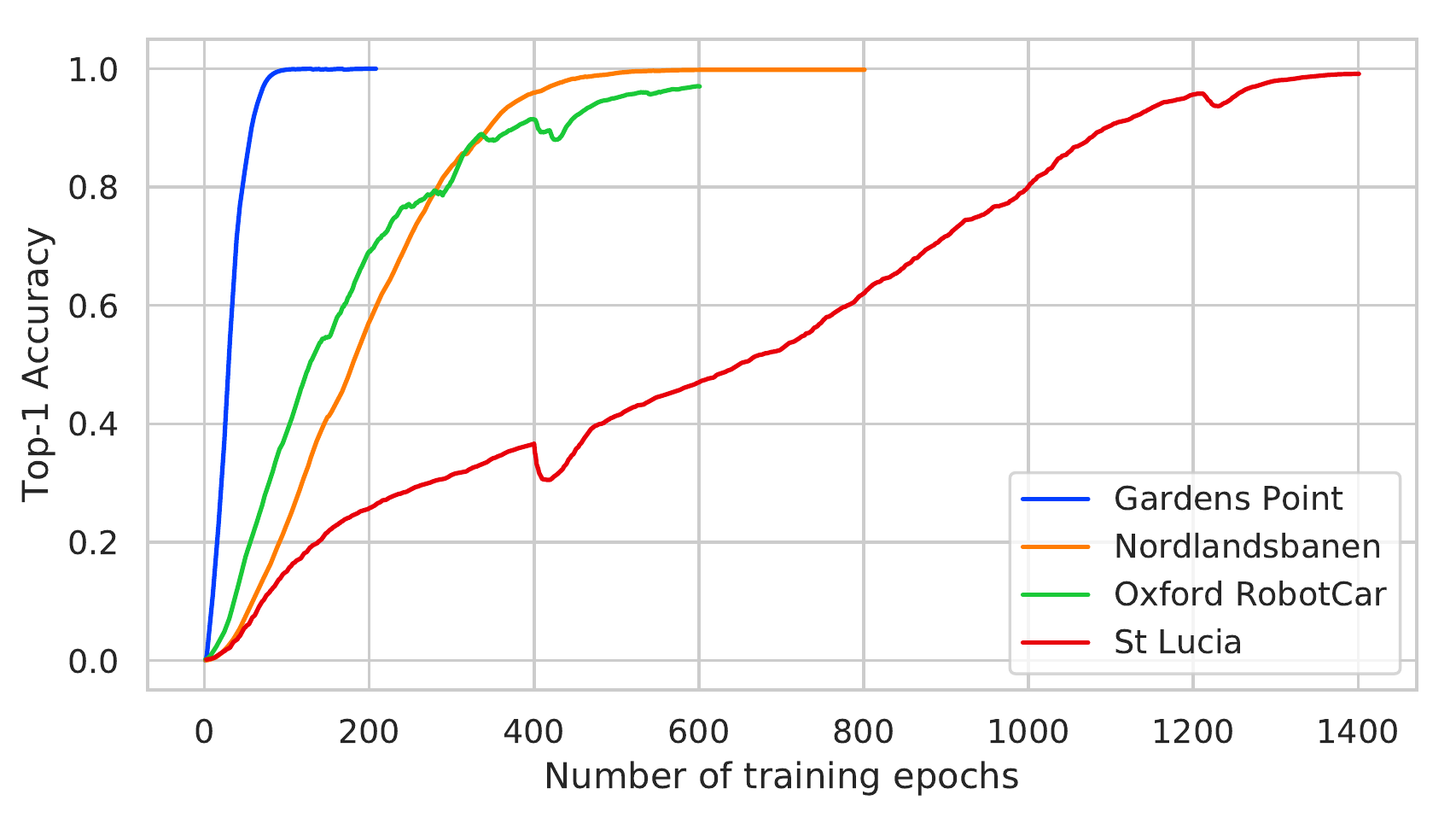}
  \includegraphics[width=0.36\linewidth]{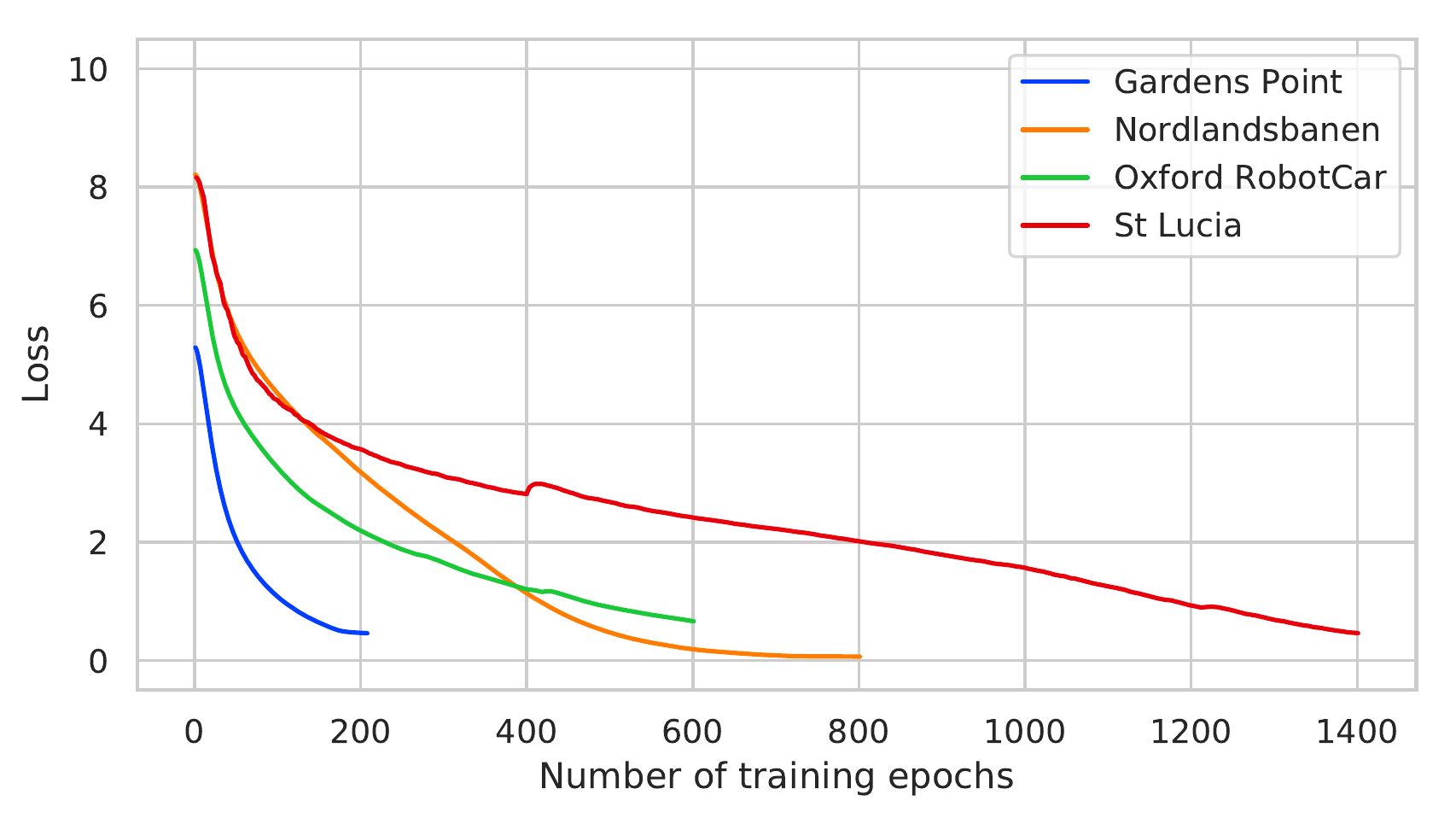}
  \caption{Top-1 accuracy and loss cost vs. training epochs.}
  \label{acc_curve}
\end{figure*}

\nopagebreak

\begin{figure*}
  \centering
  \includegraphics[width=0.32\linewidth]{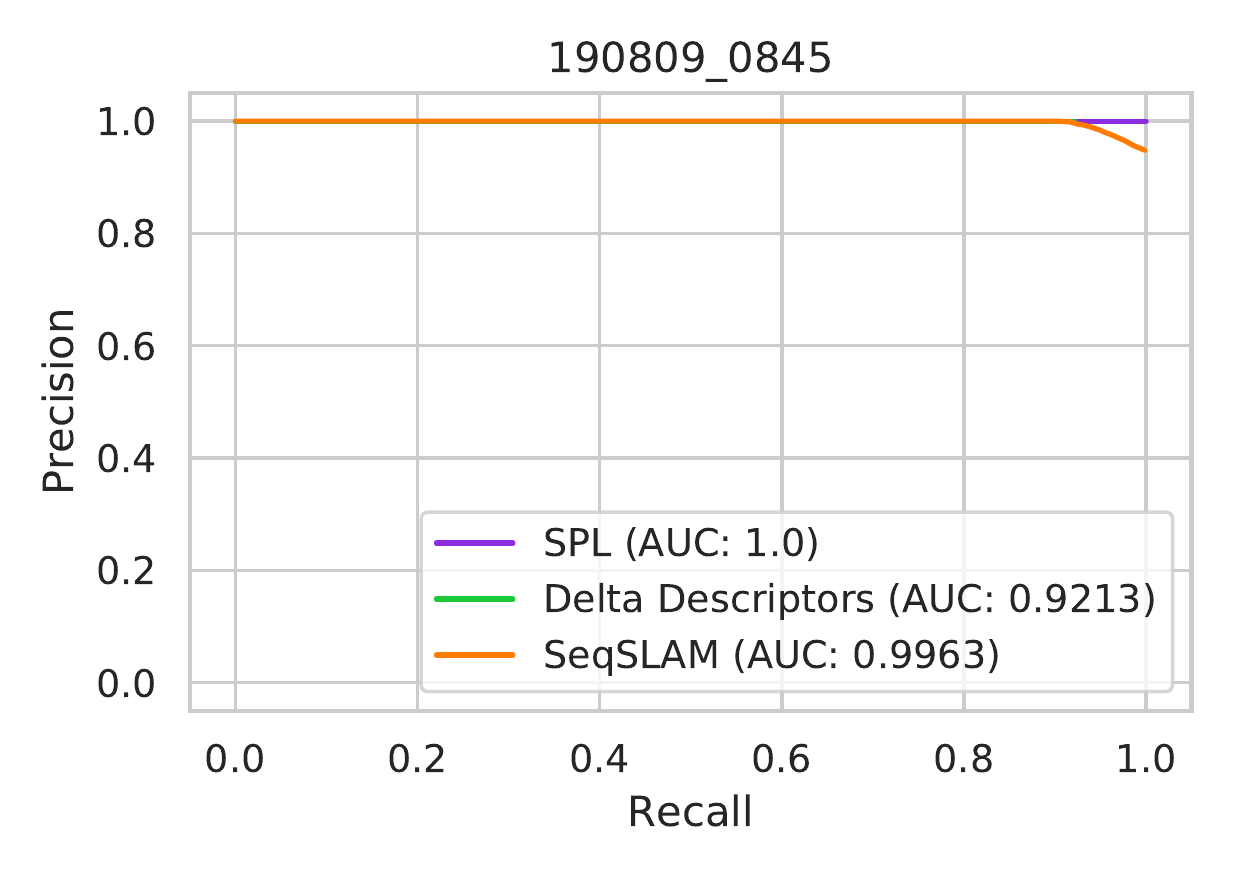}
  \includegraphics[width=0.32\linewidth]{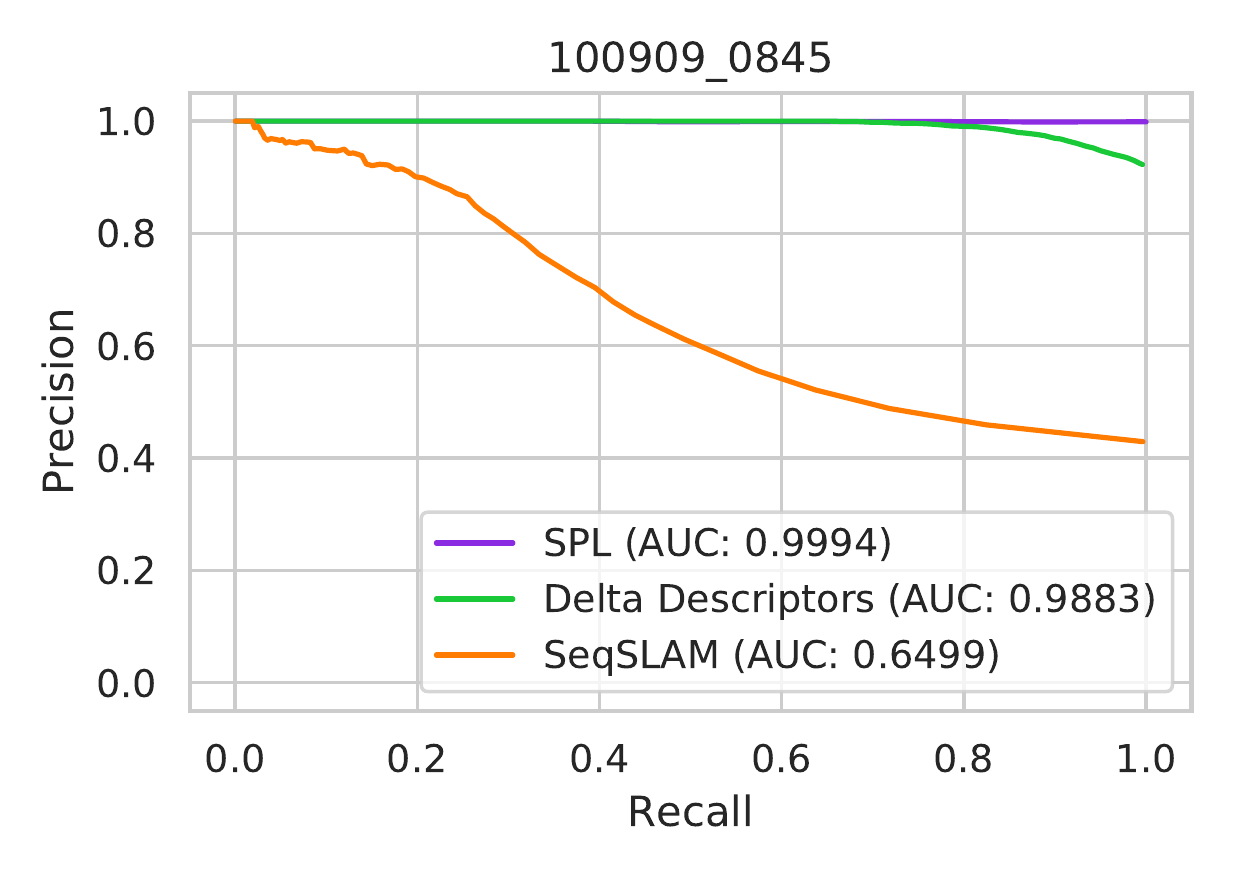}
  \includegraphics[width=0.32\linewidth]{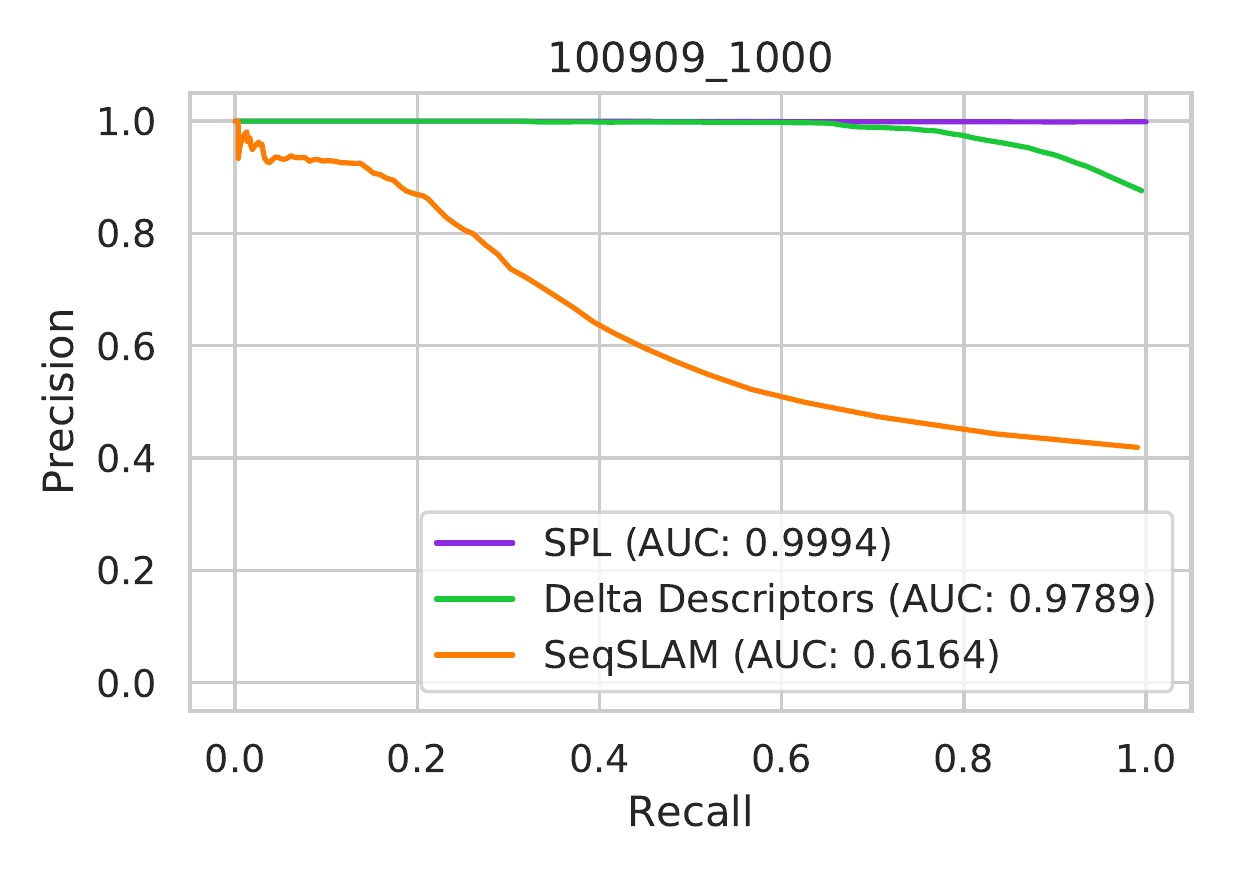}
  \includegraphics[width=0.32\linewidth]{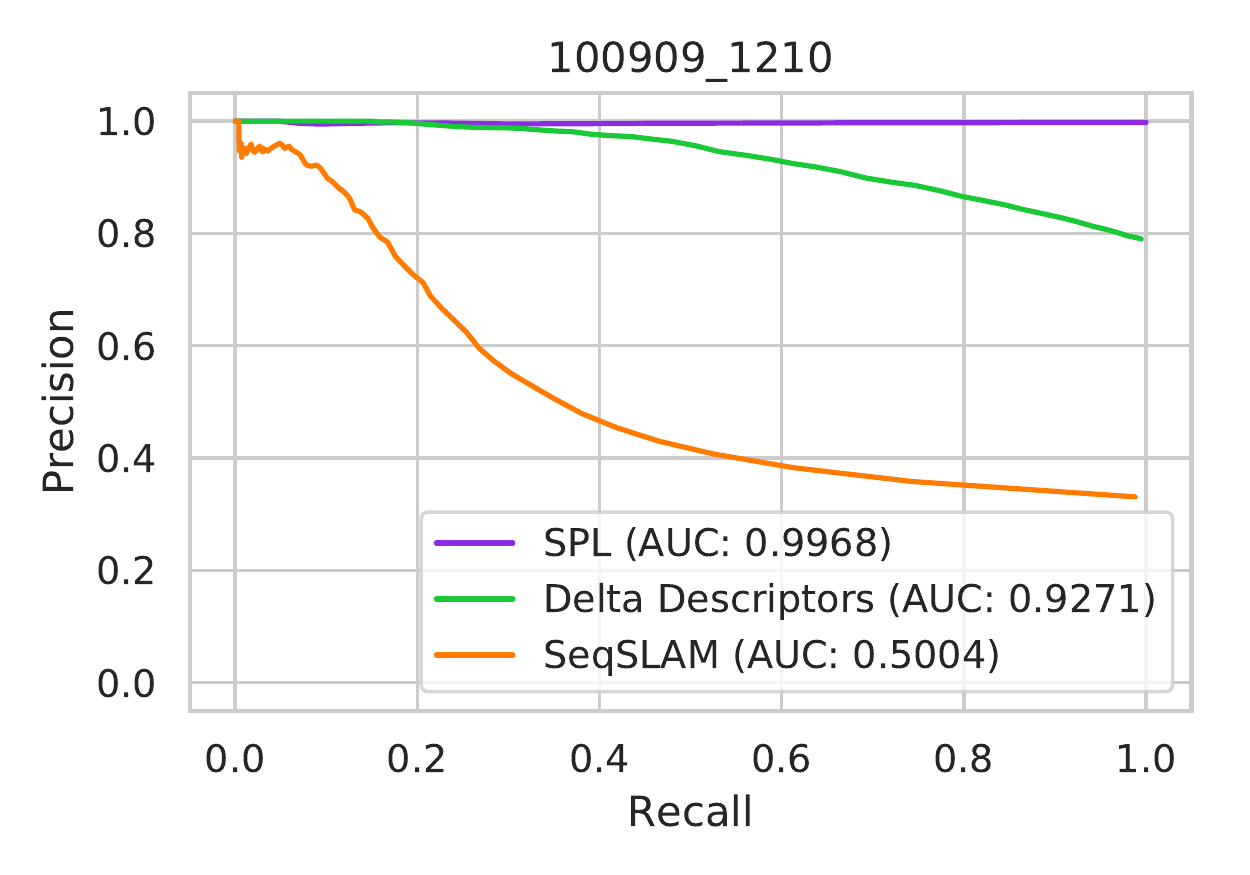}
  \includegraphics[width=0.32\linewidth]{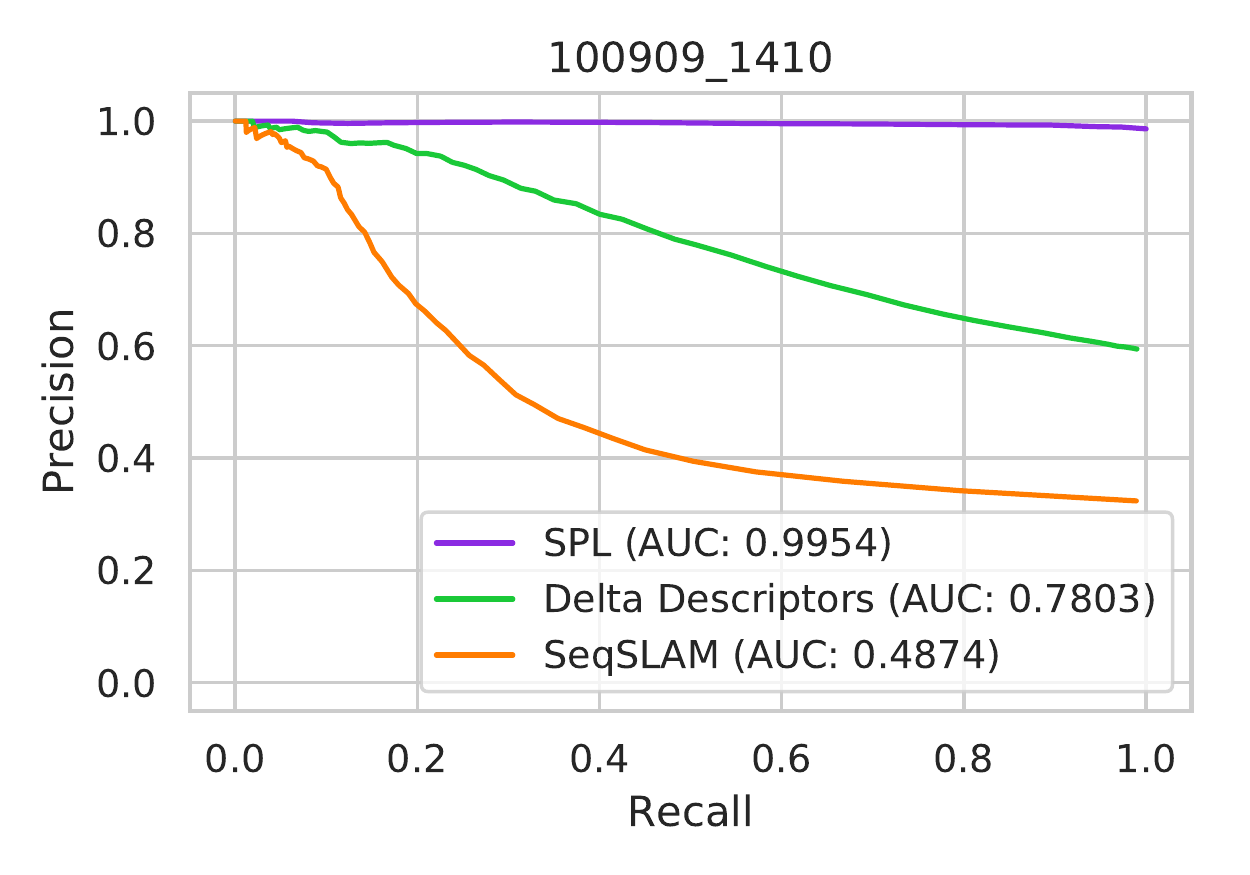}
  \includegraphics[width=0.32\linewidth]{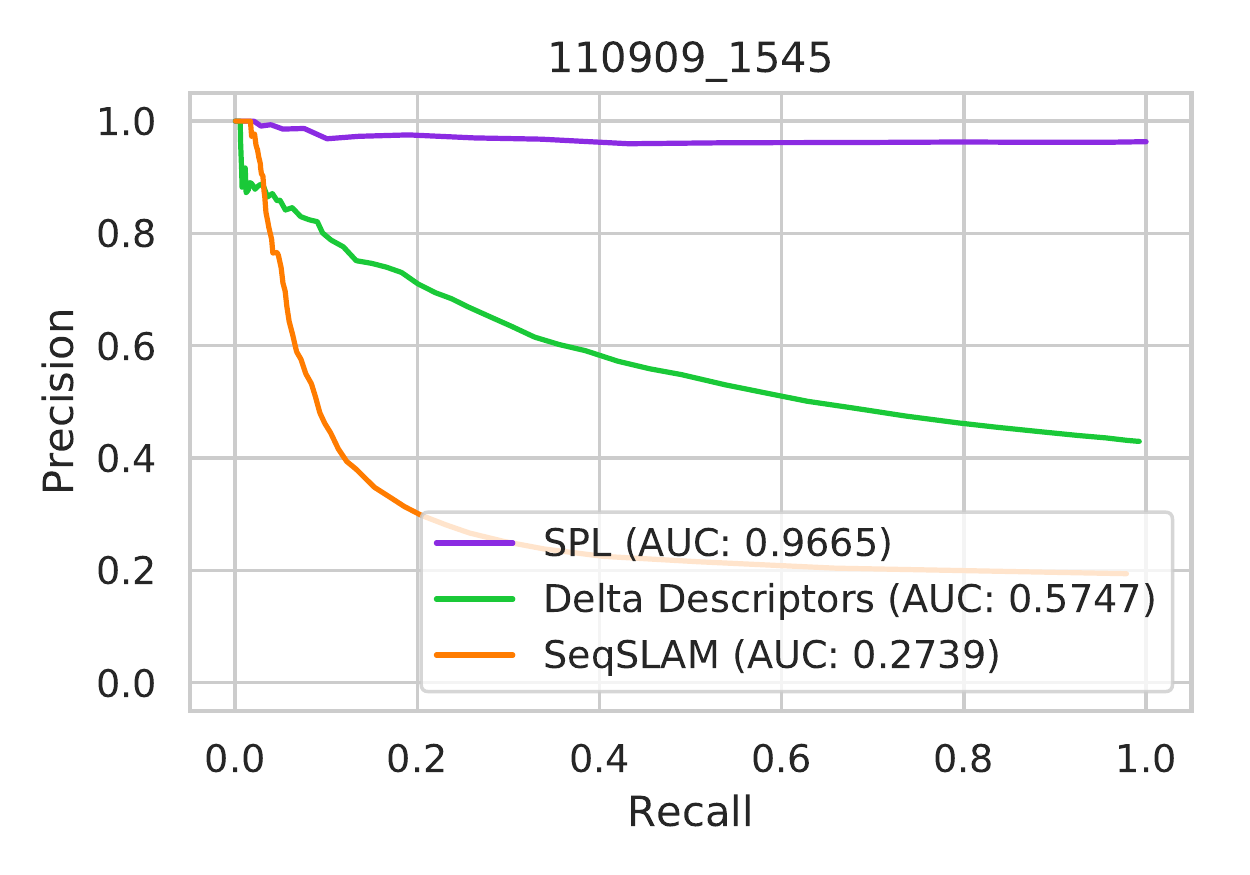}
  \includegraphics[width=0.32\linewidth]{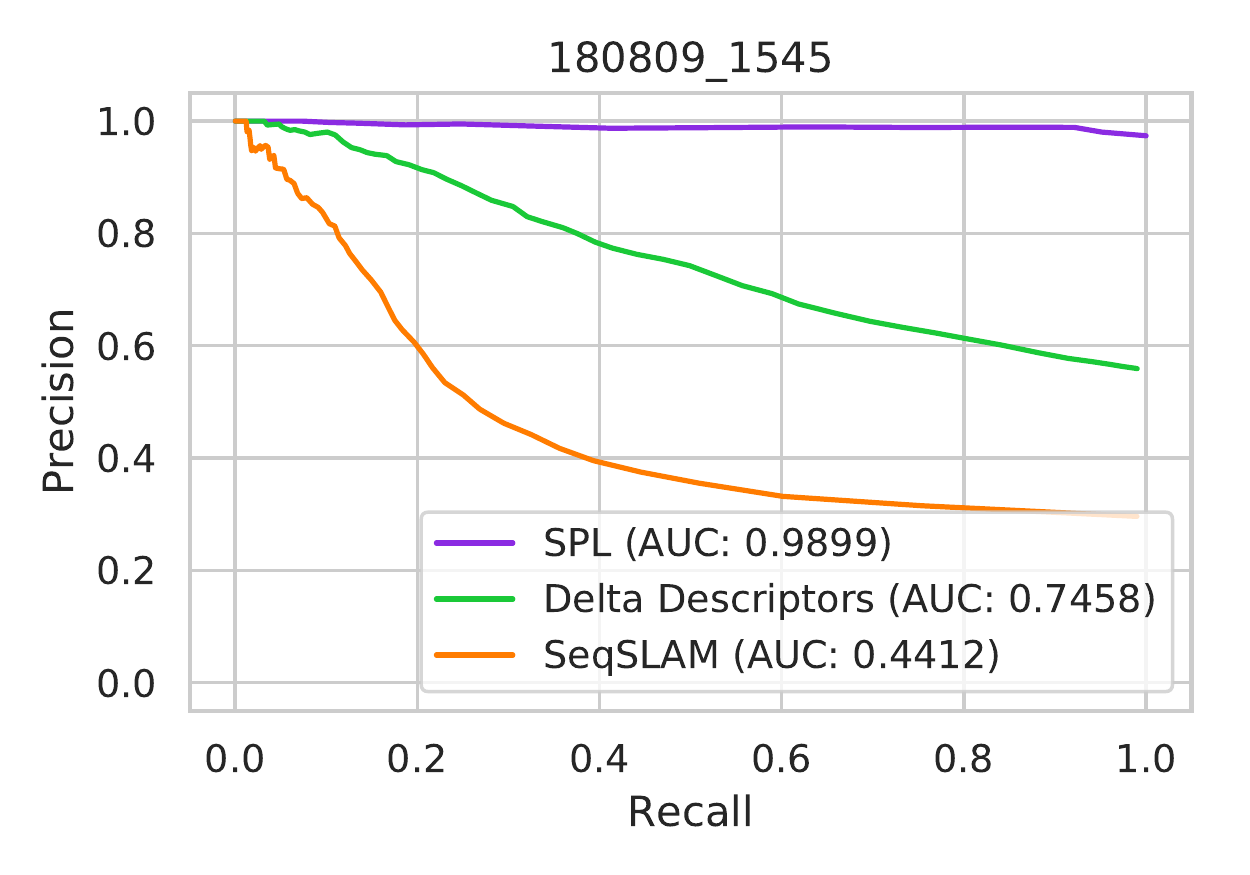}
  \includegraphics[width=0.32\linewidth]{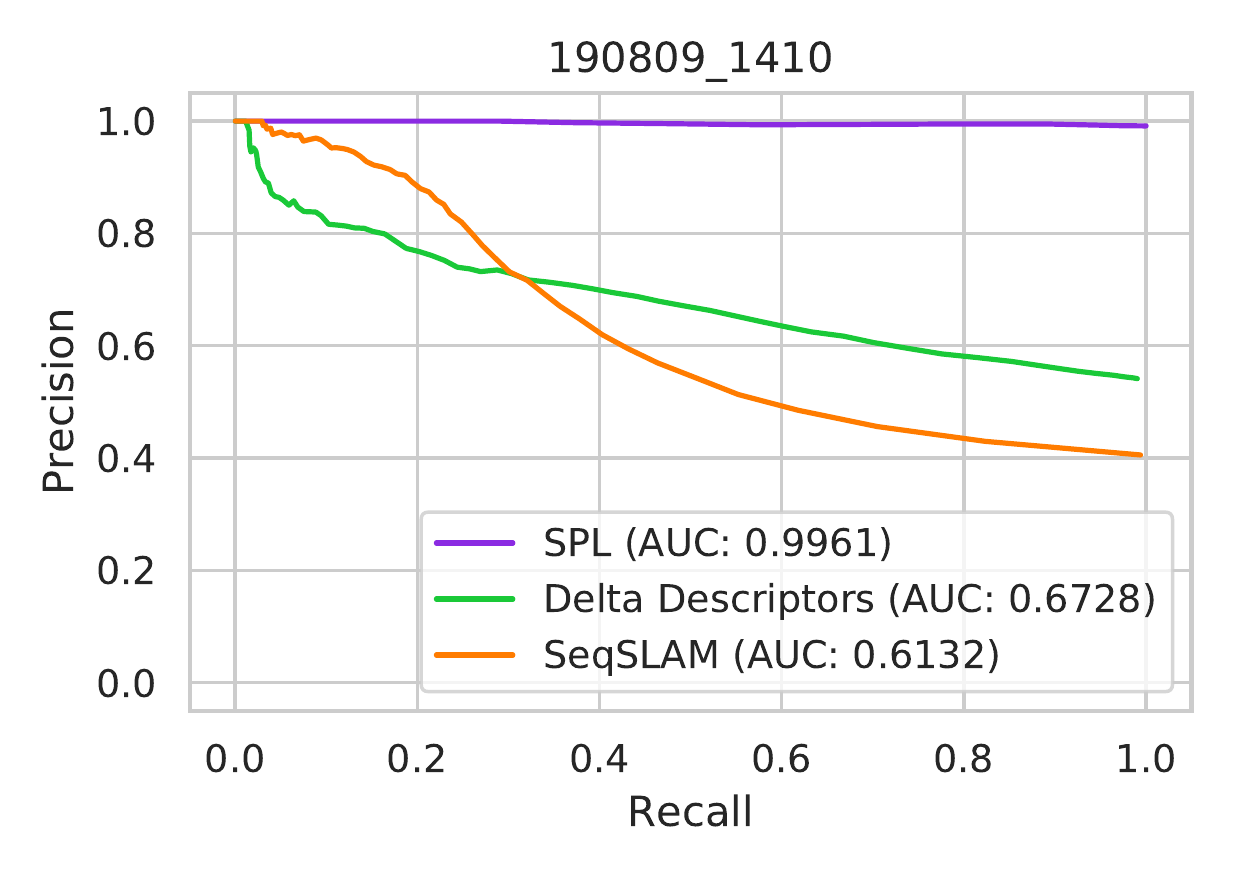}
  \includegraphics[width=0.32\linewidth]{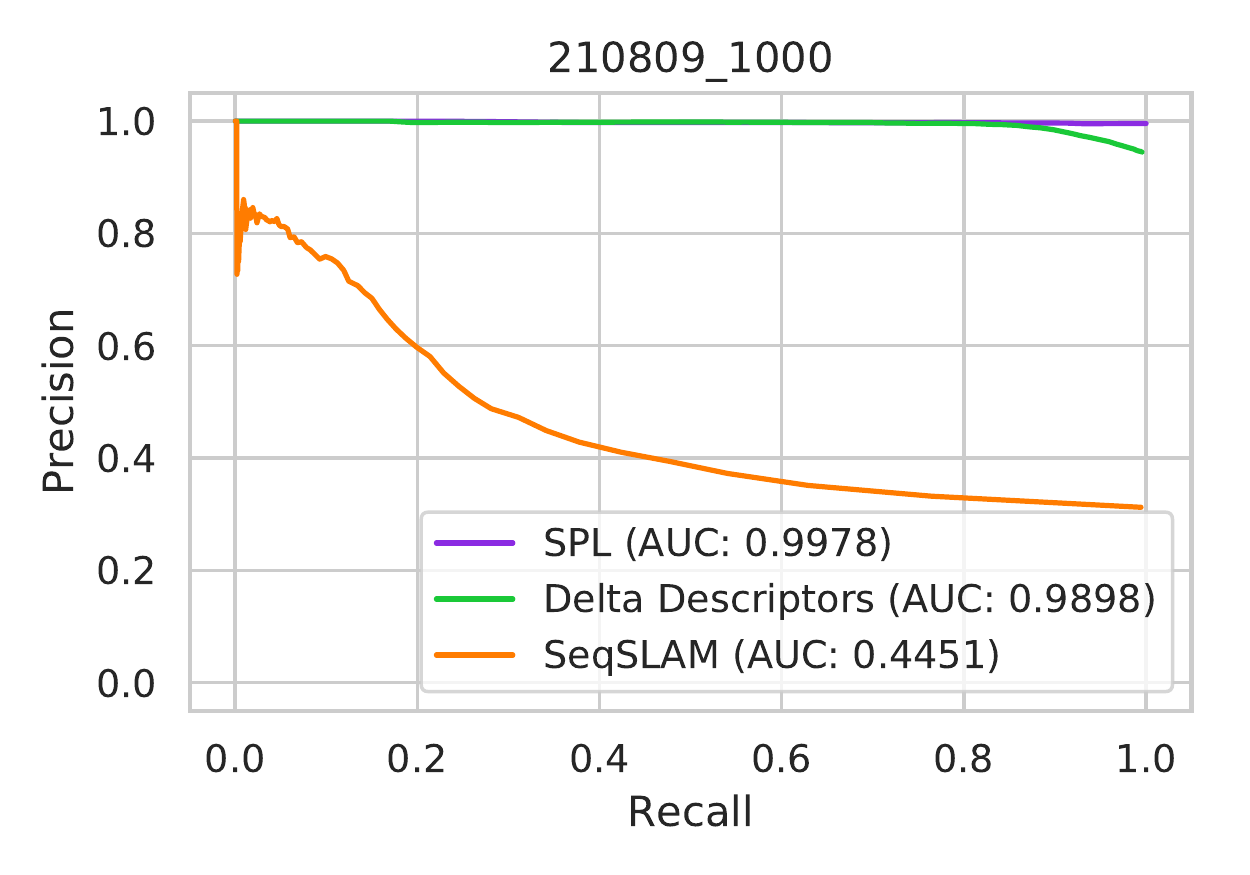}
  \includegraphics[width=0.32\linewidth]{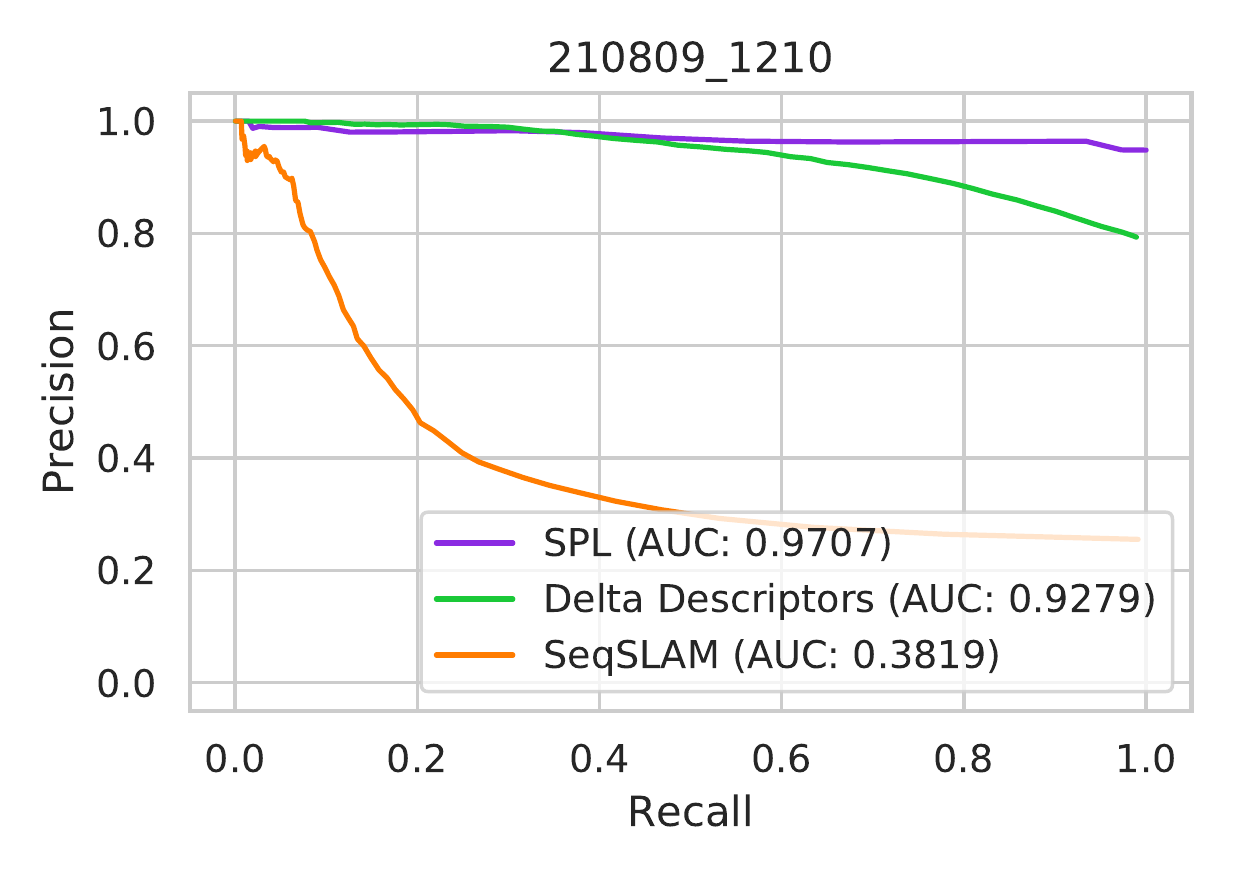}
  \vspace{-3mm}
  \caption{\textbf{PR curves using a tolerance of 20 meters at a sequence length of 10} on St Lucia. Reference traversal: 190809\_0845.}
  \vspace{-2mm}
  \label{st_pr}
\end{figure*}

\newpage

\begin{figure*}[!h]
  \centering
  \includegraphics[width=0.32\linewidth]{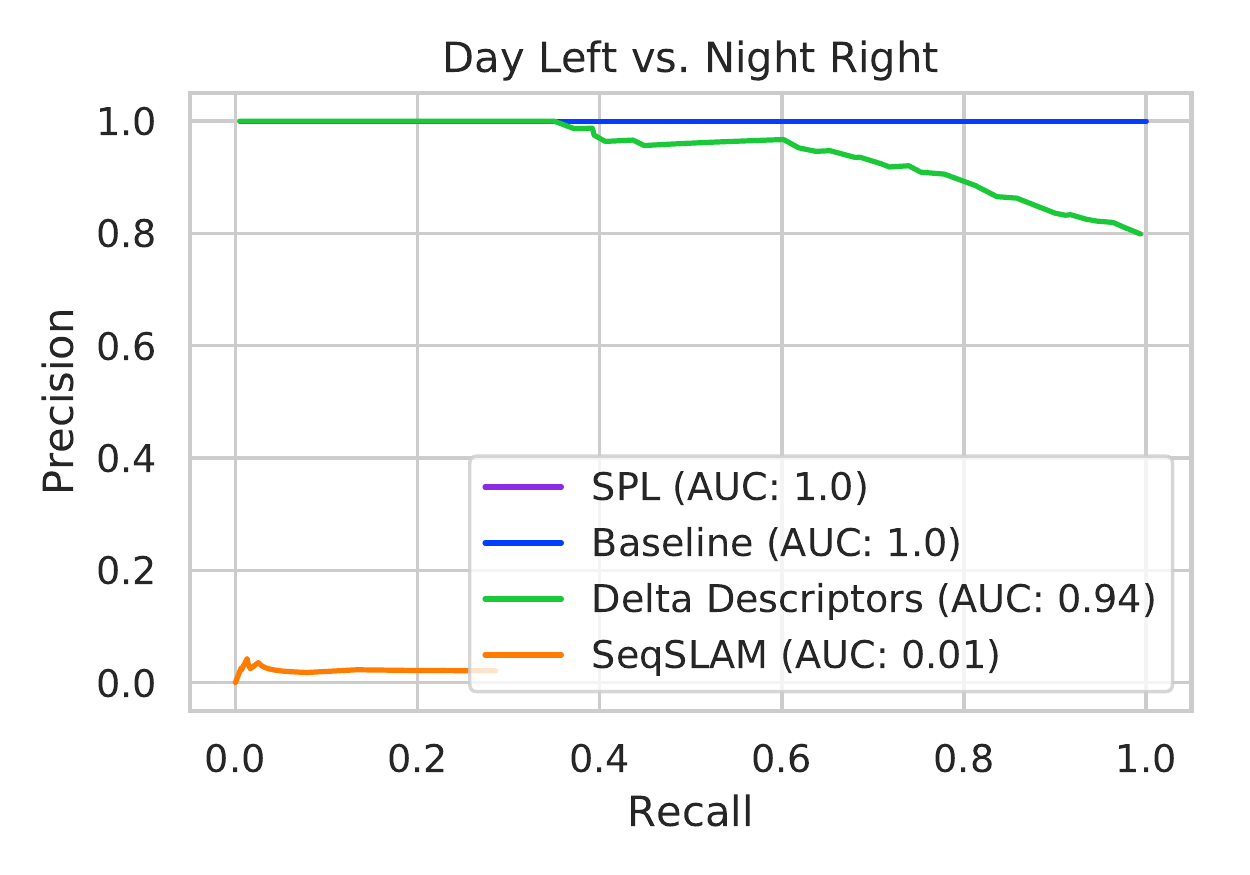}
  \includegraphics[width=0.32\linewidth]{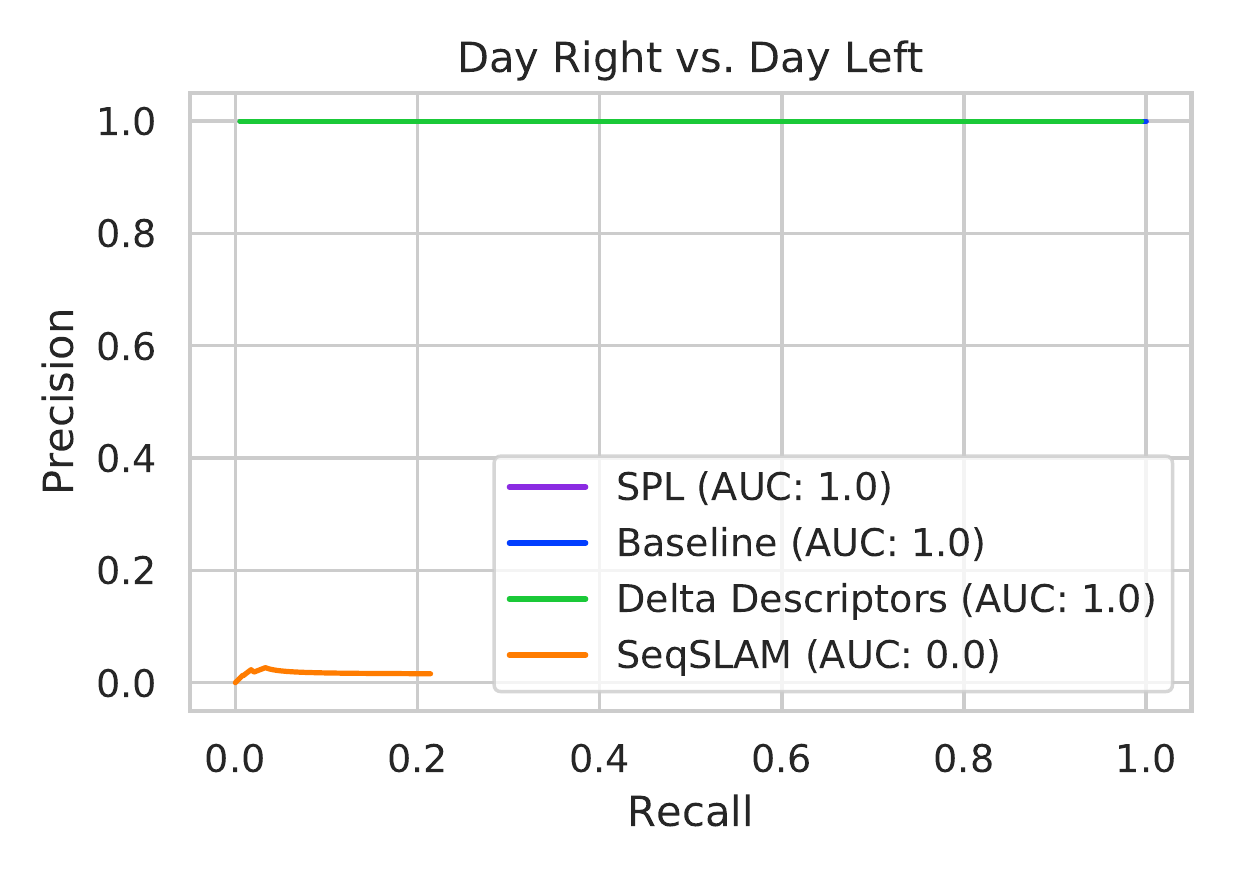}
  \includegraphics[width=0.32\linewidth]{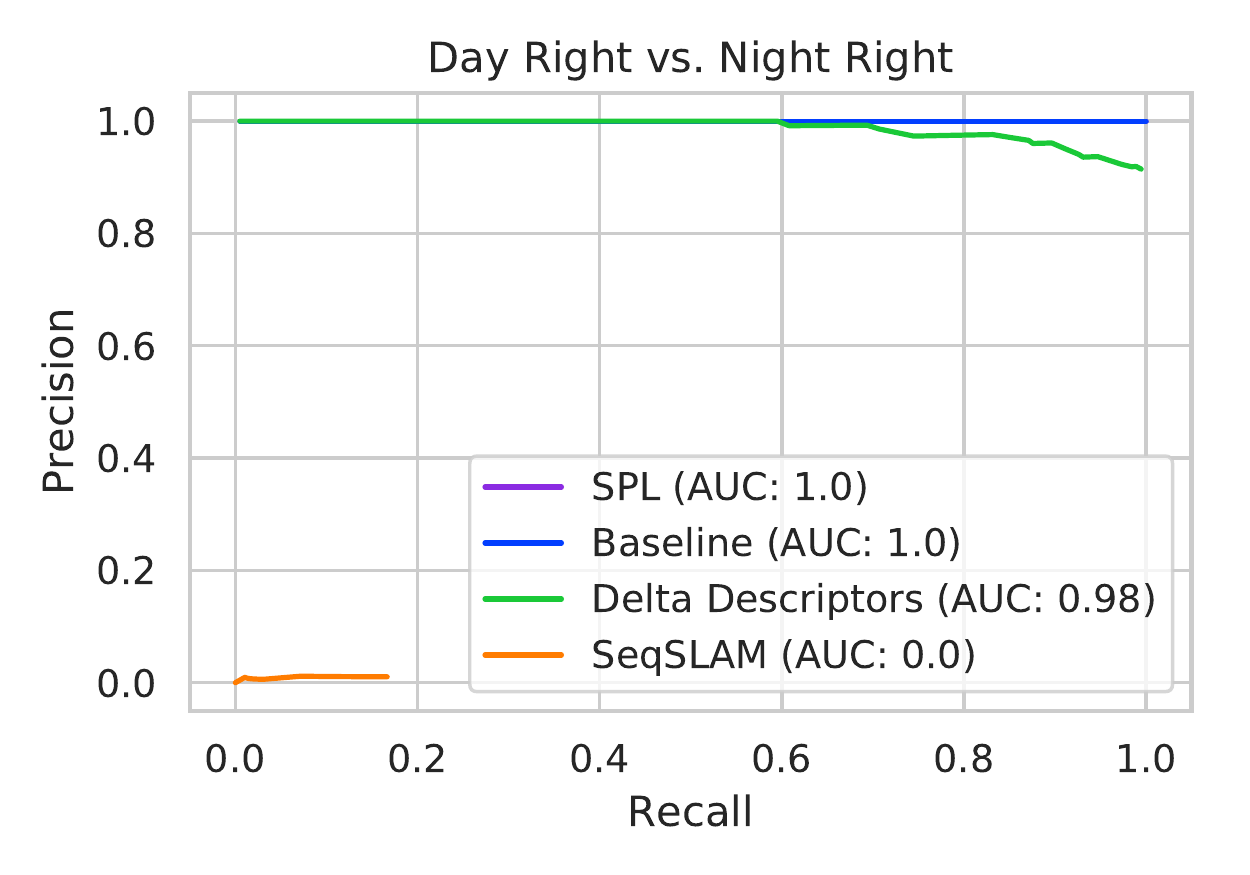}
  \vspace{-3mm}
  \caption{\textbf{PR curves using a tolerance of 10 at a sequence length of 10} on Gardens Point.}
  \vspace{-2mm}
  \label{gp_pr10}
\end{figure*}

\begin{figure*}[!h]
  \centering
  \includegraphics[width=0.32\linewidth]{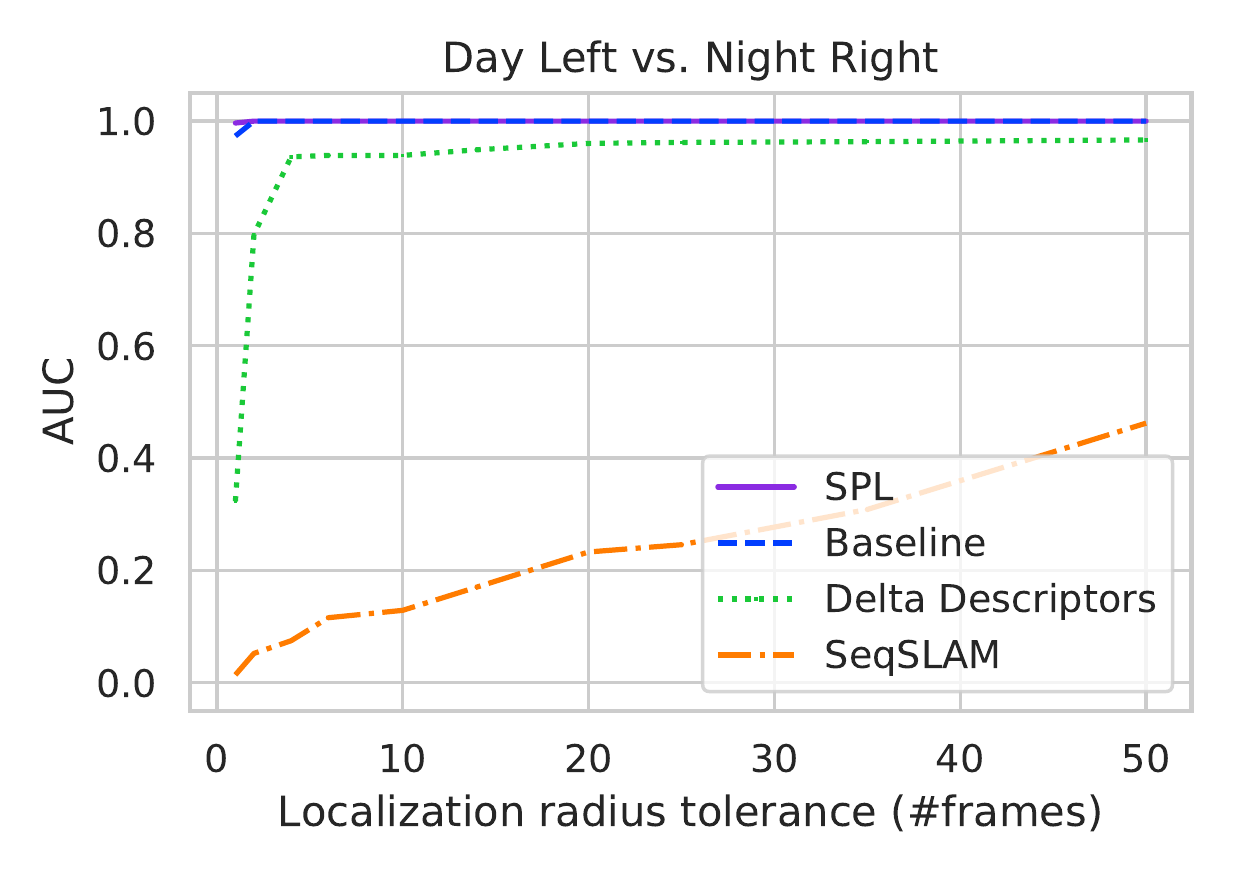}
  \includegraphics[width=0.32\linewidth]{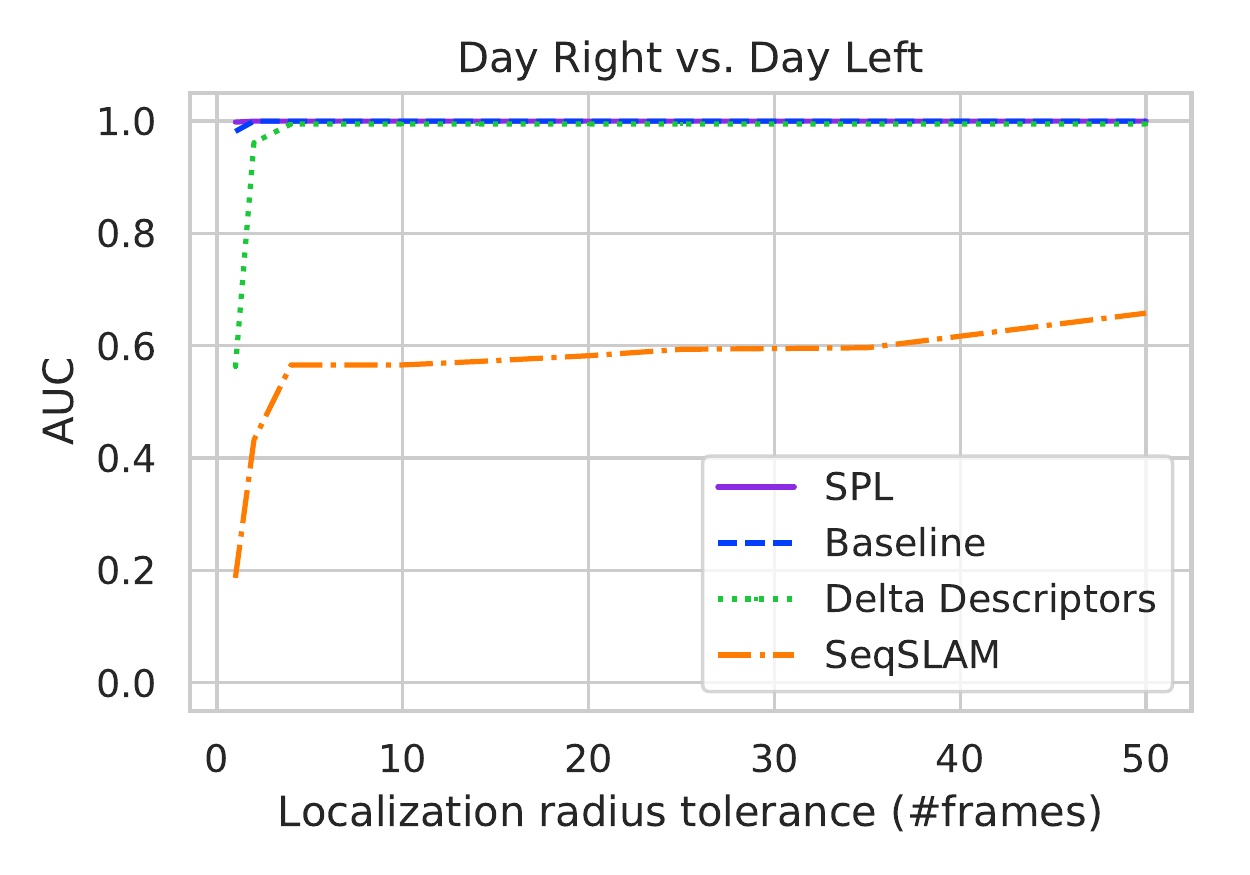}
  \includegraphics[width=0.32\linewidth]{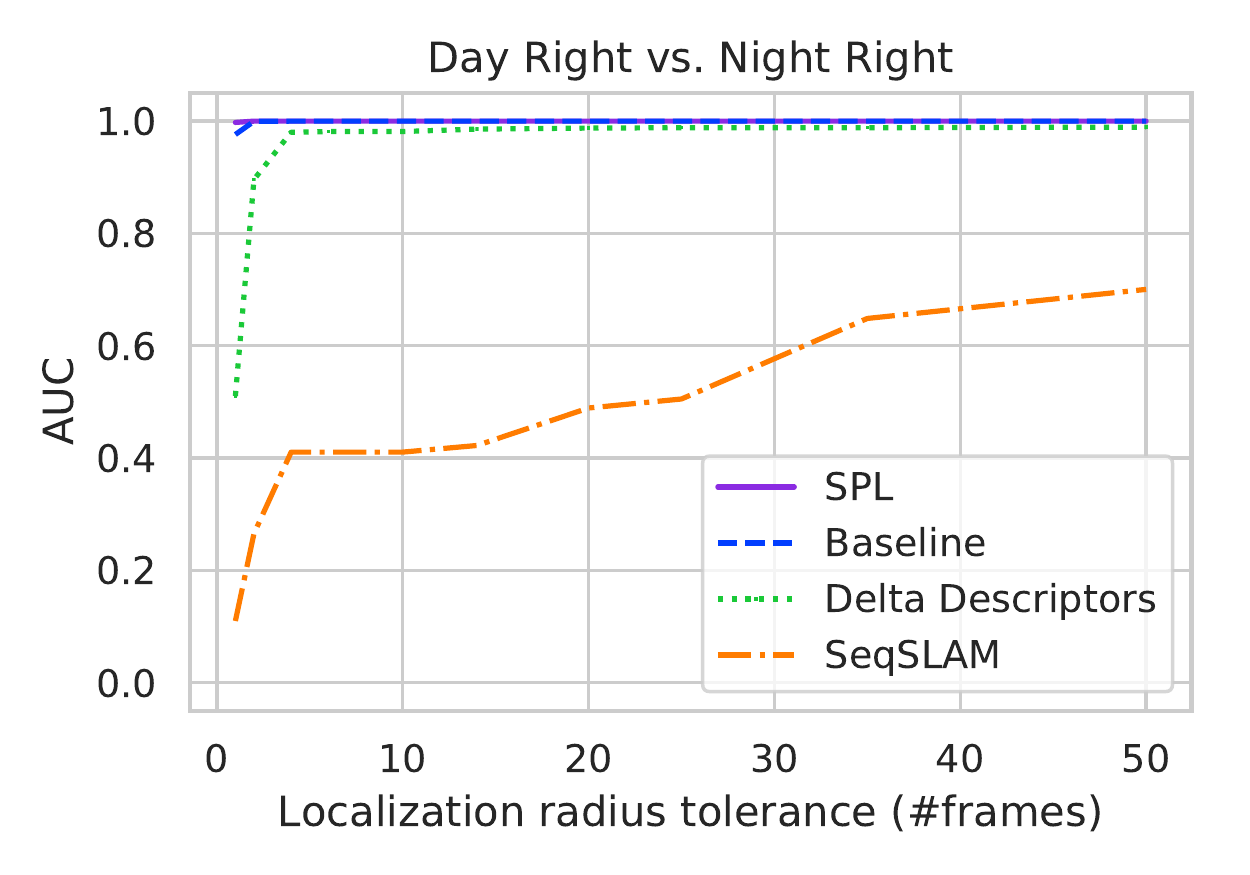}
  \vspace{-3mm}
  \caption{\textbf{AUC vs. localization tolerance at a sequence length of 10} on Gardens Points.}
  \vspace{-2mm}
  \label{gp_tol}
\end{figure*}

\begin{figure*}[!h]
  \centering
  \includegraphics[width=0.32\linewidth]{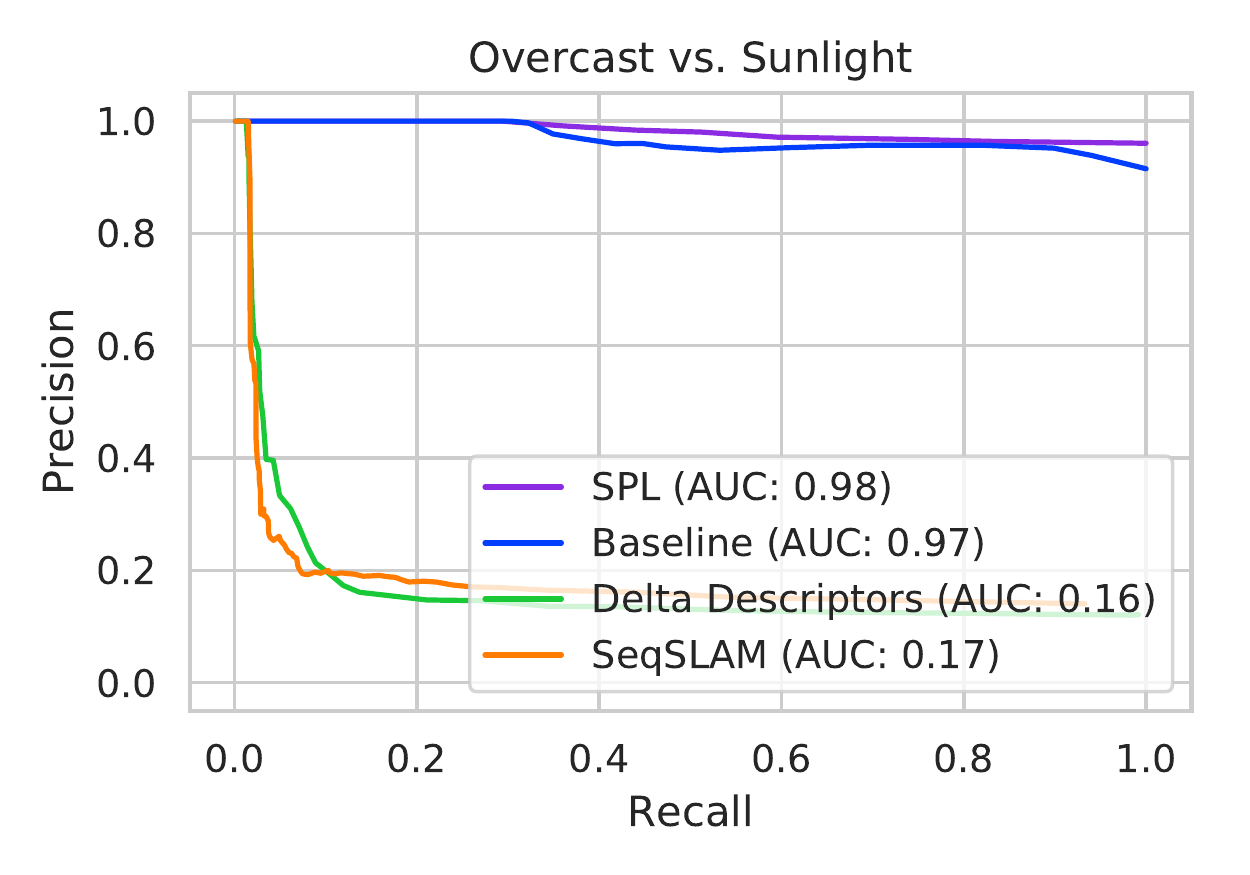}
  \includegraphics[width=0.32\linewidth]{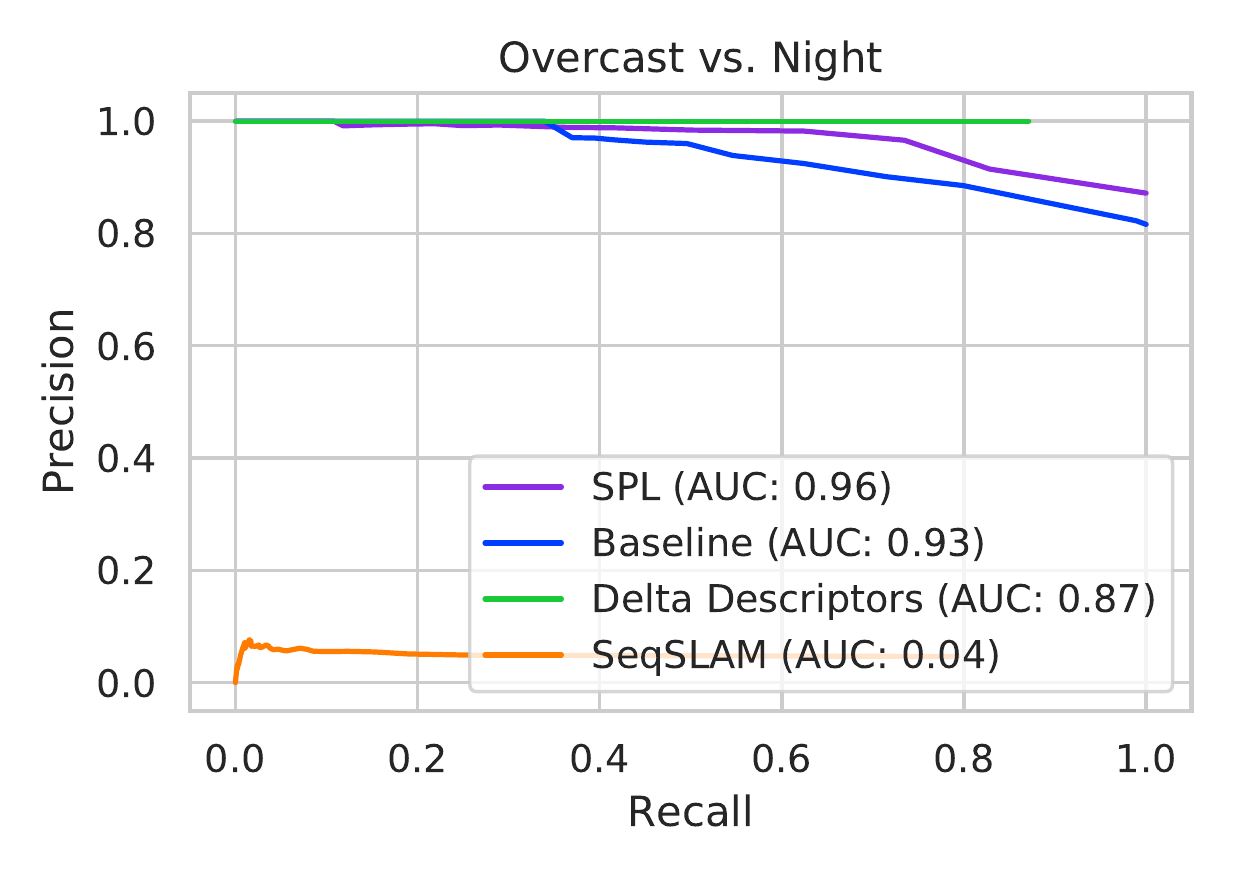}
  \includegraphics[width=0.32\linewidth]{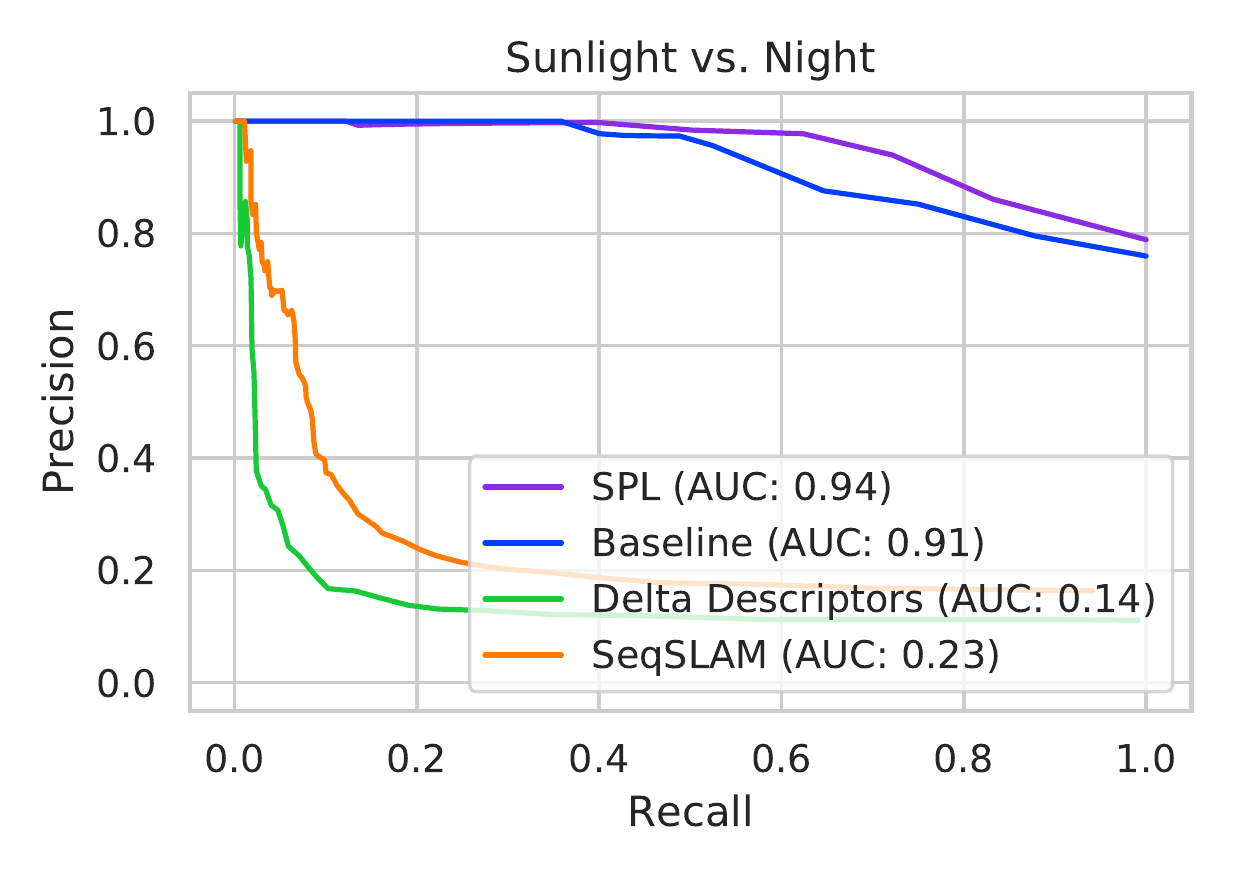}
  \vspace{-3mm}
  \caption{\textbf{PR curves using a tolerance of 10 at a sequence length of 10} on Oxford RobotCar.}
  \vspace{-2mm}
  \label{oxf_pr10}
\end{figure*}

\begin{figure*}[!h]
  \centering
  \includegraphics[width=0.32\linewidth]{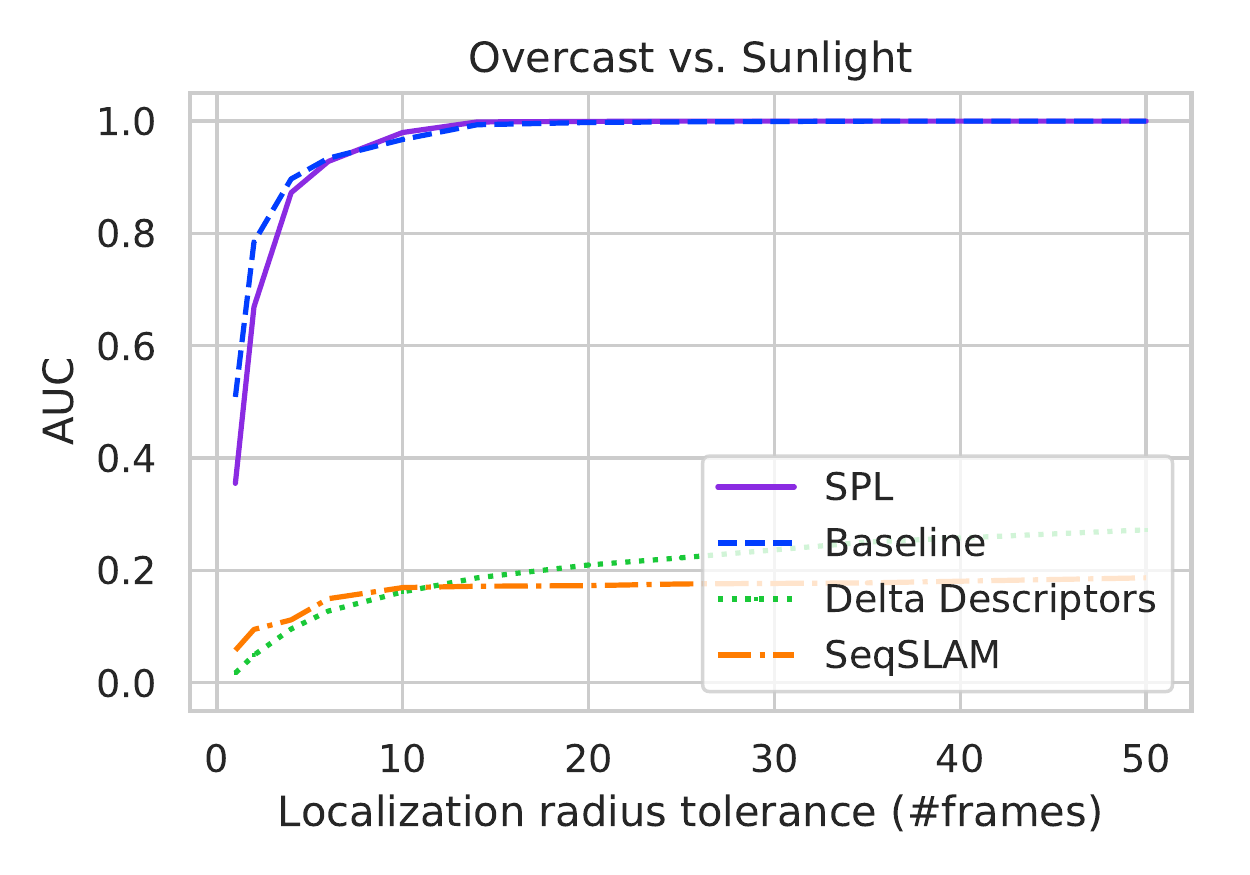}
  \includegraphics[width=0.32\linewidth]{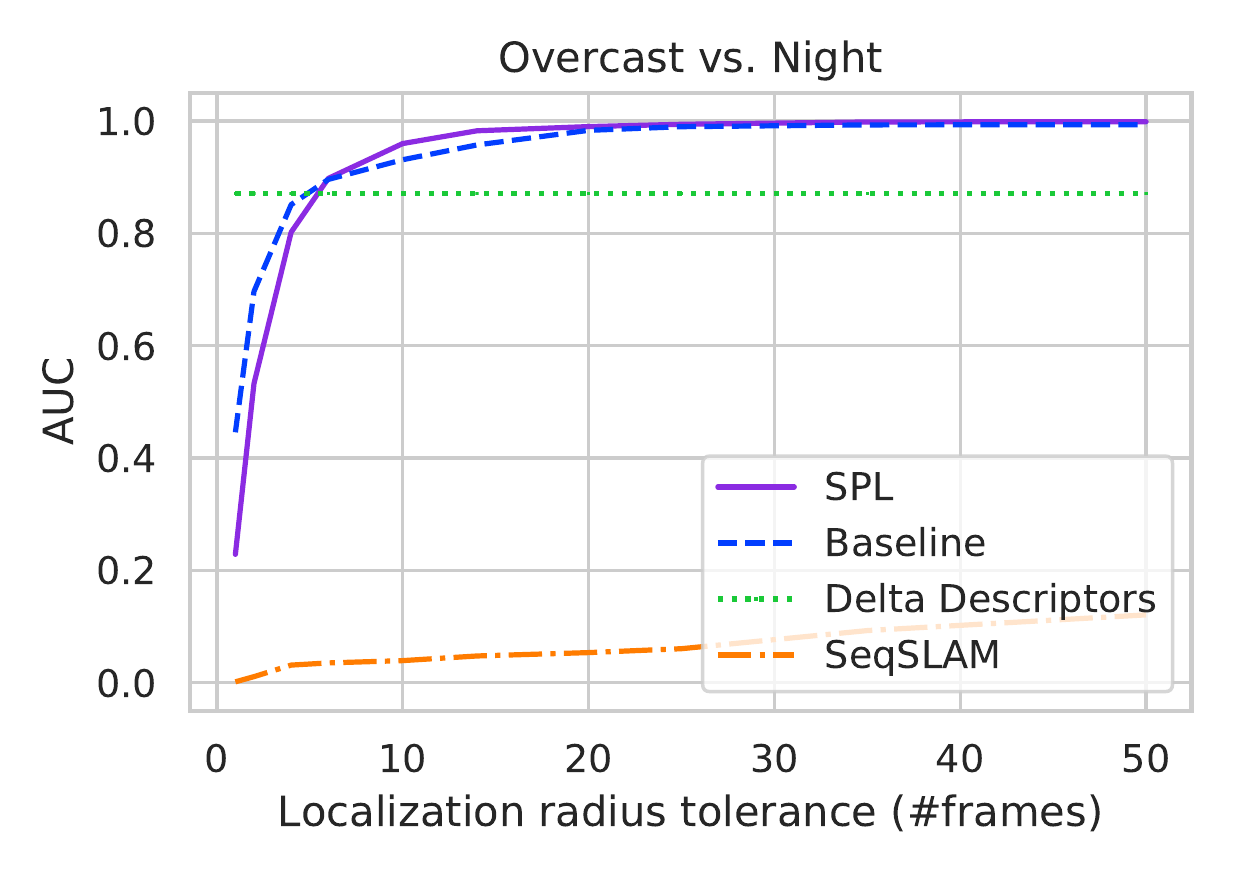}
  \includegraphics[width=0.32\linewidth]{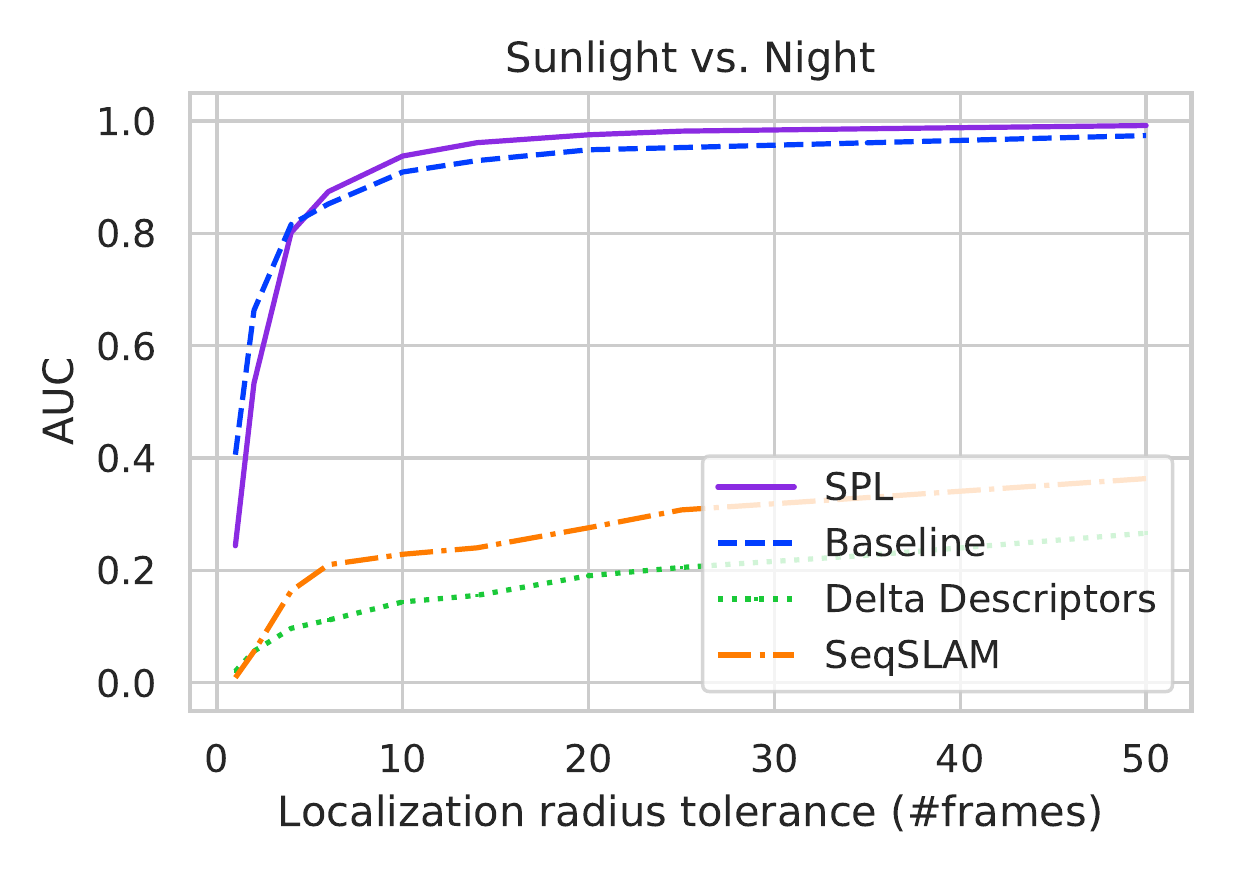}
  \vspace{-3mm}
  \caption{\textbf{AUC vs. localization tolerance at a sequence length of 10} on Oxford RobotCar.}
  \vspace{-2mm}
  \label{oxf_tol}
\end{figure*}

\begin{figure*}[!t]
  \centering
  \includegraphics[width=0.96\linewidth]{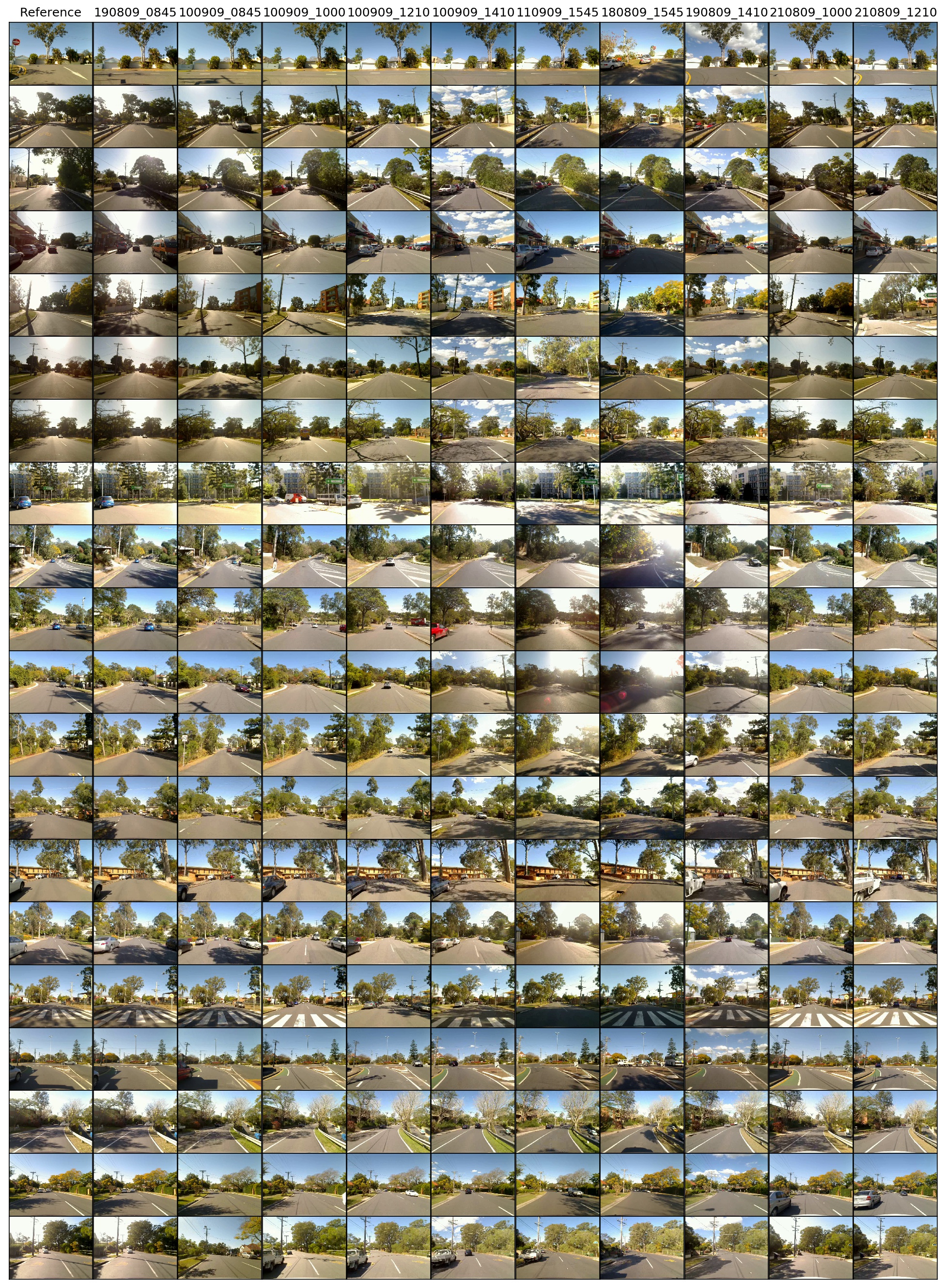}
  \vspace{-3mm}
  \caption{SPL deployment on St. Lucia. Reference traversal: 190809\_0845.}
  \vspace{-2mm}
  \label{st_qualy}
\end{figure*}

\begin{figure*}[!t]
  \centering
  \includegraphics[width=0.96\linewidth]{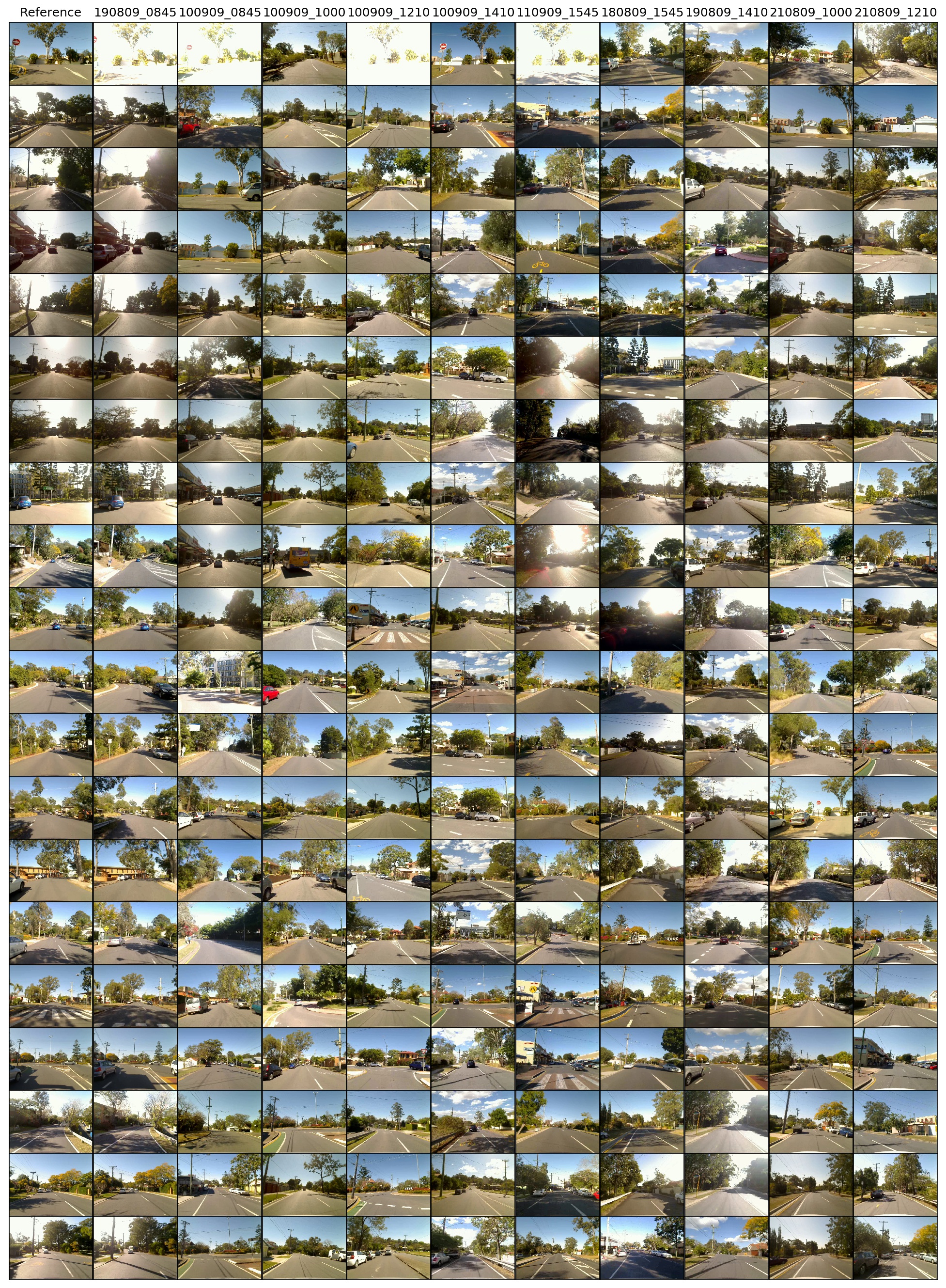}
  \vspace{-3mm}
  \caption{Delta Descriptors deployment on St Lucia. Reference traversal: 190809\_0845.}
  \vspace{-2mm}
  \label{st_dd_qualy}
\end{figure*}

\begin{figure*}[!t]
  \centering
  \includegraphics[width=0.96\linewidth]{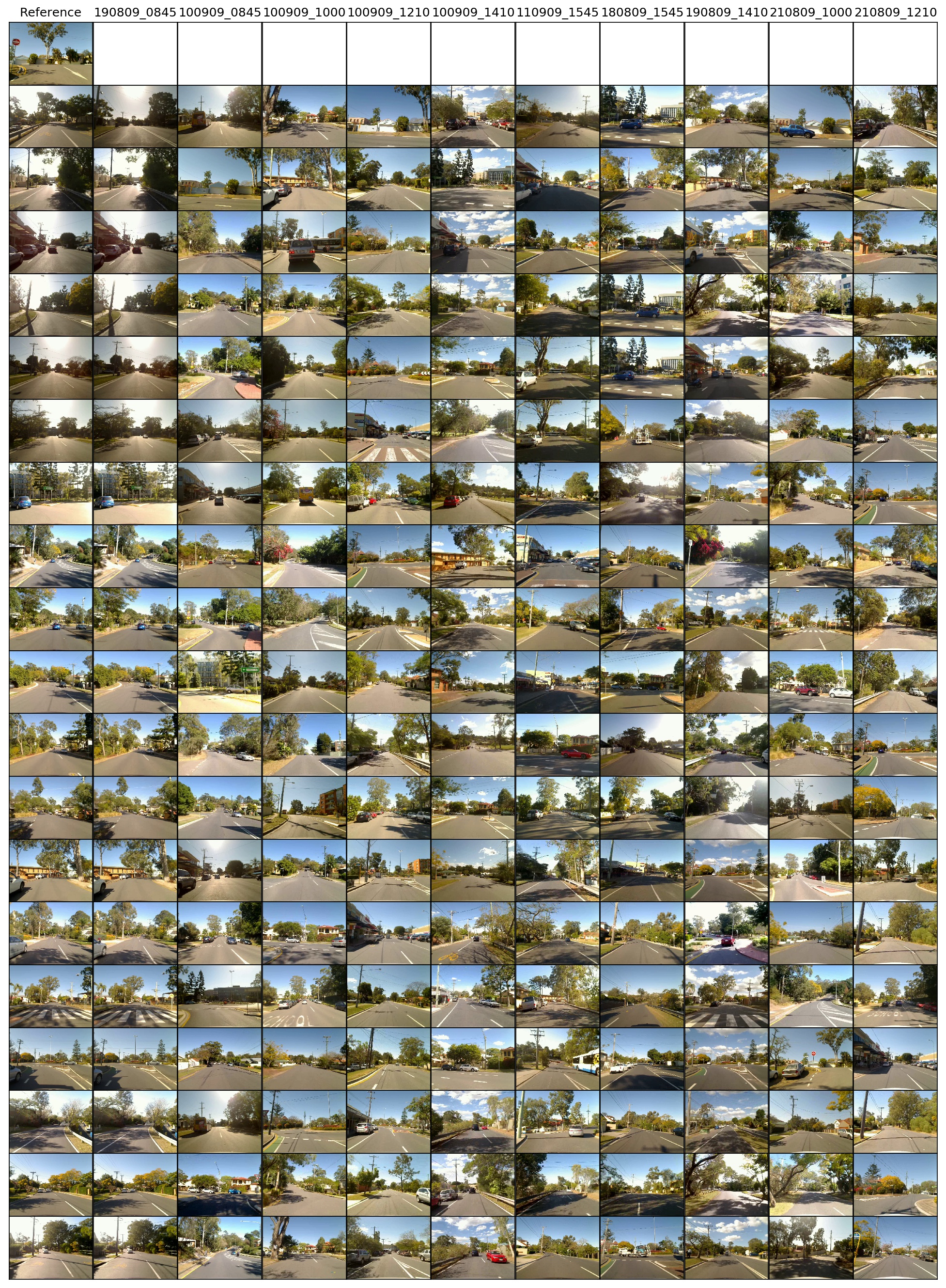}
  \vspace{-3mm}
  \caption{SeqSLAM deployment on St Lucia. Reference traversal: 190809\_0845.}
  \vspace{-2mm}
  \label{st_ss_qualy}
\end{figure*}

\end{document}